\documentclass{article}

\usepackage{iclr2027_conference,times}   %

\usepackage[hyphens]{url}
\usepackage{graphicx}
\usepackage{booktabs}
\usepackage{amsmath,amssymb}
\usepackage{xcolor}
\usepackage[hidelinks]{hyperref}
\pdftrailerid{}
\newcommand{\Q}{\lvert Q\rvert}
\newcounter{algo}   %
\newif\iflnsection
\lnsectiontrue      %
\graphicspath{{figs/}}

\title{Rank Collapse Is Recoverable,\\
Growing $\Q$ Is Not:\\
Out-of-Sample Early Warning for Value\\
Divergence in High-UTD Soft Actor-Critic}

\author{Tianqi Bu$^{1,2}$\thanks{Corresponding author: \texttt{tianqi.bu@rutgers.edu}} \quad
YuXuan Peng$^{3}$ \quad Junteng Tu$^{4}$ \quad Henghui Xiao$^{1}$ \\
$^{1}$Rutgers University \quad $^{2}$Nanjing Tech University \\
$^{3}$Nanjing University of Posts and Telecommunications \quad $^{4}$New York University\\[-18pt]}

\iclrfinalcopy

\begin{document}

\maketitle

\begin{abstract}
Raising the update-to-data (UTD) ratio breaks off-policy critics in two ways grouped as ``plasticity loss'':
collapsing representations and growing value magnitude $\Q$. We separate them in Soft Actor-Critic (SAC) with
scaled critics (width 2048, no normalization). Collapse
is survivable: at UTD ratio 16, HalfCheetah critics with most units dormant keep learning, and the training guard,
which stops runs whose loss or $\Q$ explodes, never flags them. Within one high-UTD SAC configuration, runs start
close together, and how far a critic's $\log_{10}\Q$ has climbed by step 15k, its early growth, ranks the runs by how soon
the guard flags them. At 15k, a flagged run's $\Q$ sits a median of over a hundredfold below its flag level, yet
the climb's rate already orders the flags (Harrell's C and out-of-sample AUC 0.78 on Walker2d, 0.98 on Ant, at UTD
ratio 4). Dormancy does not. Aborting on this rate saves about a tenth of
held-out Walker2d compute and stays net-positive live.\iflnsection{} A LayerNorm critic lowers the rate,
removes the flag on Walker2d at UTD ratio 4 and lowers the return.\fi
\end{abstract}

\section{Introduction}
Raising the update-to-data (UTD) ratio, the number of gradient updates per environment step (also called the replay
ratio), is a standard way to make off-policy reinforcement learning more sample-efficient
\citep{chen2021redq,hiraoka2022droq,doro2023replay}. Raised too far, it breaks scaled critics. In Walker2d at four updates per
environment step, with a critic scaled to width 2048, runs that differ only in the seed lose their critic to divergence at widely different times: some
partway through the budget, others late in it or never within it.

Two symptoms of a breaking critic share the name ``plasticity loss'' \citep{klein2024plasticity}: the
representation collapses, with falling effective rank and dormant units \citep{kumar2021implicit,sokar2023dormant},
and the value estimates grow without bound \citep{vanhasselt2018triad,hussing2024dissecting}. Practice either treats
both in every run, with resets or normalization \citep{doro2023replay,nauman2024bro,lee2025simba}, or watches the
representation, as dormancy-triggered resets do \citep{sokar2023dormant}. \citet{hussing2024dissecting} trace
high-UTD failures to value divergence. None of this work tells a practitioner, early in one run, how soon that run
will diverge.

\begin{figure}[t]\centering\includegraphics[width=\linewidth]{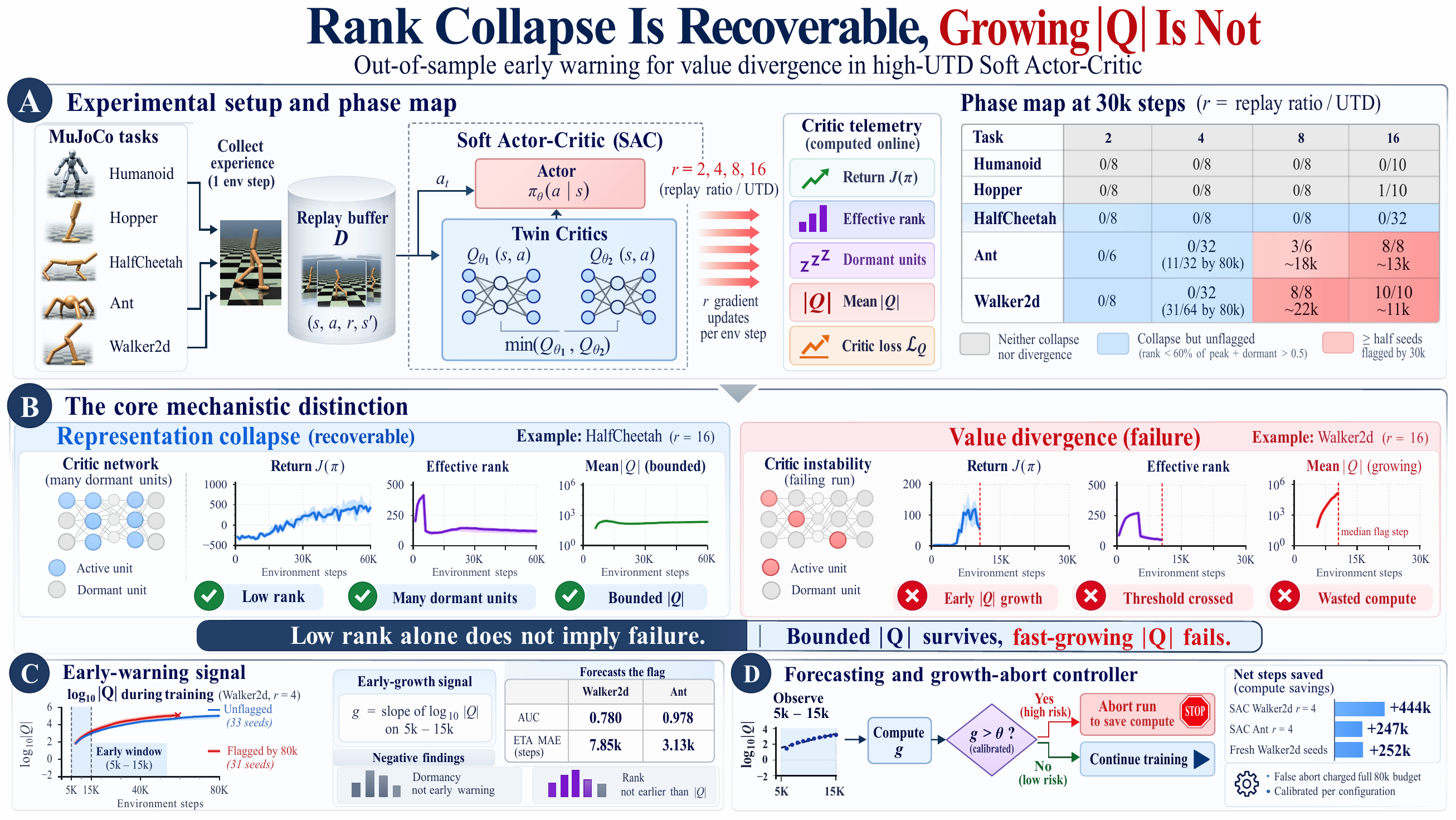}
\caption{\textbf{Within one high-UTD SAC configuration, runs start close together, and how far a critic's $\log_{10}\Q$ has
climbed by step 15k, its early growth, ranks the runs by how soon the guard flags them.}
(A)~Setup, critic telemetry, and the phase map at 30k steps (flagged/seeds per cell).
(B)~A collapsed HalfCheetah critic at $r{=}16$ keeps $\Q$ bounded and keeps learning; Walker2d critics at $r{=}16$
grow $\Q$ until the guard flags them (dashed: median flag step).
(C)~$\log_{10}\Q$ of the Walker2d $r{=}4$ seeds the guard flags by 80k and of those it leaves unflagged (median and
interquartile range); $g$ is read in the shaded early window. Inset: AUC for a flag by 80k (one cut of the order) and flag-step error
(Table~\ref{tab:forecast}).
(D)~One run's early-window $\Q$ and its slope $g$; the growth-abort controller aborts runs whose $g$ exceeds the
abort threshold $\theta$. Bars: net steps saved (Table~\ref{tab:ctrl}a).}\label{fig:overview}
\end{figure}

A training guard stops a run once its critic loss or the critic's value magnitude $\Q$ explodes; we say that it
flags the run. We show that a critic can lose rank at learning onset and most of its units to dormancy, and still
keep learning. HalfCheetah critics at UTD ratio 16 do so, and the guard flags none of their
runs. Its flags follow fast growth of $\Q$ instead (Figure~\ref{fig:overview}B). The \emph{post-learning rank drop},
the fall of rank from its peak after learning starts, co-emerges with that growth: it starts at about the same
time. Within one high-UTD SAC configuration, runs start close together, and how far a critic's $\log_{10}\Q$ has climbed by
step 15k, its early growth, ranks the runs by how soon the guard flags them. We call the rate of this
climb the \emph{early-growth signal} $g$. A minimal model, in which each seed sets its own rate on a shared growth
profile, is consistent with faster early growth bringing an earlier flag (Section~\ref{sec:mechanism}); at a fixed
early level the fitted sign runs opposite to its prediction (Appendix~\ref{app:nested}). The
\emph{growth-abort controller} aborts
the runs whose $g$ exceeds a calibrated abort threshold.

Our evidence is 612 training runs of Soft Actor-Critic (SAC; \citealp{haarnoja2018sac}) and Twin Delayed Deep
Deterministic policy gradient (TD3; \citealp{fujimoto2018td3}). They cover five MuJoCo tasks at four UTD ratios,
all with one unnormalized critic of width 2048\iflnsection{}; Section~\ref{sec:ln} adds 32 seed pairs that compare it
with a LayerNorm critic\fi. %
In Walker2d and Ant at UTD ratio 4, where the seeds split, $g$ orders the runs by time to the flag with Harrell's
concordance index C of 0.78 and 0.98, and it reaches the same leave-one-seed-out area under the receiver operating
characteristic (ROC) curve (AUC) for a flag within the budget. The growth-abort controller saves 9.9\% of the
Walker2d cohort's compute out of sample and stays net-positive live. Code and per-seed logs are available from the corresponding author on request.

In the title's sense, collapse is recoverable: the return of the collapsed critic still rises
(Section~\ref{sec:decouple}). Fast
$\Q$ growth is not: it marks the runs that reach the guard's loss
threshold within the budget. The trainer computes $\Q$ in every critic update, so the warning needs
no extra forward pass.
\begin{itemize}\setlength{\itemsep}{1pt}\setlength{\parskip}{0pt}
\item \textbf{Collapse is survivable; fast $\Q$ growth marks the flagged runs} (Section~\ref{sec:decouple}). At UTD
ratio 16, HalfCheetah critics with more than three quarters of their units dormant keep learning,
and a periodic reset ends ahead at 60k on all 6 paired seeds. Dormancy carries no seed-level warning of a flag, and the
post-learning rank drop co-emerges with $\Q$ growth.
\item \textbf{Early $\Q$ growth ranks runs by how soon the guard flags them} (Section~\ref{sec:forecast}), per
seed and out of sample, in the two SAC configurations whose seeds split, and it ranks the soonest flags best. It is
no threshold extrapolation: at 15k a flagged run's $\Q$ is still a median of over a hundredfold short of its flag
level, and over half of the unflagged Walker2d runs end at or above the lowest flag level.
\item \textbf{A shared-profile model is consistent with the direction of the flag order} (Section~\ref{sec:mechanism}). Runs
start close together and flag in a narrow band of $\Q$, so faster early growth goes with an earlier flag in every
cohort we test, more loosely on Walker2d than the model implies. The climb bends, so the timing needs a fitted map.
\item \textbf{The growth-abort controller saves compute out of sample and live} (Section~\ref{sec:control}), with
every false abort charged the full budget. It saves more than abort rules that watch dormancy or return. We calibrate its abort threshold once per configuration.
\iflnsection
\item \textbf{A LayerNorm critic removes the flag on Walker2d at UTD ratio 4}
(Section~\ref{sec:ln}). It lowers the rate $g$ and ends at a lower median return.
\fi
\end{itemize}

\section{Related Work}\label{sec:related}
\paragraph{Loss of plasticity.} Networks trained on changing targets lose plasticity through implicit
under-parameterization \citep{kumar2021implicit}, dormant neurons \citep{sokar2023dormant}, primacy bias
\citep{nikishin2022primacy} and capacity loss \citep{lyle2022capacity,lyle2023understanding}. The same loss appears
in on-policy agents \citep{juliani2024plasticity} and in continual learning \citep{dohare2024loss,abbas2023loss}.
Recent work ties it to a collapsing Hessian spectrum \citep{prakash2026spectral} and to churn
\citep{tang2025churn}, and keeps a proximal policy optimization actor plastic with resets and distillation
\citep{zhou2025sustainable}. Much of this line treats collapse as the failure. Two findings point elsewhere: in
offline reinforcement learning, representation rank tracks performance poorly \citep{gulcehre2022implicit}, and
\citet{lyle2024disentangling} name large regression targets as one of several separate causes of plasticity loss.
We build on both: we separate collapse from value growth online and per seed, find dormancy and the rank drop at
learning onset survivable, and forecast each run's flag from its $\Q$ growth.

\paragraph{Value divergence.} Temporal-difference learning with function approximation and bootstrapping, the
``deadly triad'', can drive values up without bound \citep{vanhasselt2018triad,achiam2019characterizing}. SEEM
predicts from a kernel analysis whether value estimates diverge in offline reinforcement learning, and the order of
their growth \citep{yue2023seem}. At high UTD,
\citet{hussing2024dissecting} trace failures blamed on overfitting early data to value divergence, as we find.
\citet{nauman2024bitter} compare over sixty regularized agents, \citet{rybkin2025scales} set the UTD per
configuration from scaling fits, and \citet{bjorck2022variance} trace outlier seeds to early numerical instability
after the fact and propose fixes that reduce it. These analyses explain divergence per configuration or after the
fact. We rank the seeds at step 15k, out of sample, by how soon the guard will flag them, and measure what acting on
the ranking saves in environment steps.

\paragraph{Stabilizing high-UTD training.} Resets \citep{nikishin2022primacy,doro2023replay,nikishin2023injection},
critic normalization and regularization, and model-generated data act on every run to remove or delay the
instability. The critic methods include LayerNorm \citep{ba2016layernorm}, spectral normalization \citep{bjorck2021deeper},
regularization toward the initial weights \citep{skumar2023l2init}, the batch normalization of CrossQ
\citep{bhatt2024crossq} and its weight-normalized extension \citep{palenicek2025scaling}, BRO \citep{nauman2024bro},
and SimBa and SimbaV2 \citep{lee2025simba,lee2025simbav2}. Others add sparsity \citep{ma2025sparsity} or gradient
control \citep{creus2025stablegradients}, and MAD-TD adds model-generated data \citep{voelcker2025madtd}. We
instead forecast the instability per seed where it still occurs\iflnsection{}. On Walker2d at $r=4$, a LayerNorm
critic removes the flag (Section~\ref{sec:ln}).\else{}; whether the forecast
holds under these methods is open (Section~\ref{sec:limits}).\fi

\paragraph{Early stopping and early warning.} Hyperparameter search stops runs early on low return, with the median
stopping rule \citep{golovin2017vizier} or successive halving \citep{li2018hyperband,li2020asha}.
\citet{wortsman2023proxies} forecast Transformer instabilities from growing logit and attention norms, the
supervised analogue of watching $\Q$. Early-warning signals of transitions are standard for dynamical systems
\citep{scheffer2009earlywarning,dakos2012methods,dylewsky2023universal,deb2024optimal,li2026tipping}. Section~\ref{sec:control} scores the growth-abort controller against the
median stopping rule. We model the flag step as a right-censored event
\citep{kaplan1958nonparametric,cox1972regression,chen2024survival}.

\section{Setup}\label{sec:setup}
\paragraph{Agent and tasks.} We train SAC with twin critics, a tanh-Gaussian actor and automatic entropy tuning.
Actor and critics have two hidden layers of width 2048 and no normalization, with batch 2048. This is a deliberate
stress test of scaled critics, the lever BRO and SimBa pull, without their normalization,
where divergence is common enough to study seed by seed; standard $256\times2$ networks with batch 256 keep $\Q$ bounded on
Walker2d at $r=16$ over 40k steps (Appendix~\ref{app:stdwidth})\iflnsection{}, and Section~\ref{sec:ln} shows what a LayerNorm critic buys\fi. We apply no
gradient clipping outside a separate clipping test (Appendix~\ref{app:mech}). The UTD
ratio is $r\in\{2,4,8,16\}$, and learning starts after a 5k-step warm-up. The tasks are HalfCheetah, Walker2d,
Hopper, Ant and Humanoid (v5) in Gymnasium \citep{towers2024gymnasium} on MuJoCo \citep{todorov2012mujoco}, with one
hyperparameter set (Appendix~\ref{app:impl}).

\paragraph{Telemetry.} In each telemetry record we log the return, the critic loss and $\Q$ (the batch-mean
$|Q_1(s,a)|$ of the first critic over the record's gradient steps). On a held-out probe batch we measure
the effective rank \citep{roy2007effrank} and the dormant fraction \citep{sokar2023dormant} of the first critic's
penultimate layer (Appendix~\ref{app:impl}).

\paragraph{The divergence label.} The training guard stops a run at the first gradient step at which the critic
loss exceeds $10^8$, the batch-mean $|Q_1|$ exceeds $10^6$, or either becomes non-finite. We then call the run
\emph{flagged} at its flag step $T_{\rm flag}$. With the training budget as horizon $H$, we call a run flagged by $H$
if $T_{\rm flag}\le H$, and we treat a run still unflagged at $H$ as right-censored at $H$. In every unclipped
flagged run the loss criterion fired first, while $\Q$ was of order $10^5$ (Appendix~\ref{app:labels}).

\paragraph{Collapse.} We call a (task, $r$) configuration \emph{collapsed} when the median over its seeds of each
run's final-to-peak effective-rank ratio is below 0.6 and the median final dormant fraction exceeds 0.5. In most
phase-map runs the peak is the untrained critic's (Appendix~\ref{app:collapse}), so the rule includes the rank drop
at learning onset. We also measure the post-learning rank drop, the fall from the largest rank after learning starts.

\paragraph{The early-growth signal.} The early-growth signal $g$ is the least-squares slope of $\log_{10}\Q$ against
environment steps over the records in $(5\text{k},15\text{k}]$, the first 10k steps after the warm-up. We read it
at the decision step $t_d=15$k. We fixed this window once, on the first Walker2d replicate, before we ran the second replicate, the Ant cohort and the live seeds.

\paragraph{Evaluation.} We evaluate forecasts leave-one-seed-out (LOSO) within one configuration: a logistic
regression fit on the other seeds scores the held-out seed, and AUC intervals are percentile bootstraps over seeds.
We test forecasts in the \emph{split configurations}, Walker2d and Ant at $r=4$. They are the two SAC configurations in
which the guard flags some seeds of an 80k cohort and leaves the others unflagged (Appendix~\ref{app:split}).
Table~\ref{tab:design} lists the cohorts the
results use and their roles.

\begin{table}[t]\centering\footnotesize
\caption{\textbf{Experiment design (SAC).} $H$: horizon; ``Flagged'': runs the guard flags by $H$. Role:
\emph{fit/eval}, abort thresholds and classifiers fit on some seeds and scored on held-out ones; \emph{eval}, never
used for fitting; \emph{char.}, characterization. Appendix~\ref{app:compute} lists the TD3 and further
cohorts.}
\label{tab:design}
\setlength{\tabcolsep}{3pt}
\begin{tabular}{@{}lp{5.4cm}lccl@{}}
\toprule
Cohort & Tasks (UTD ratio $r$) & Seeds & $H$ & Flagged & Role \\
\midrule
Phase map & all 5 tasks, $r\in\{2,4,8,16\}$ (Fig.~\ref{fig:overview}A) & 6--32 each & 30k & 30/234 & char. \\
Forecast & Walker2d ($r{=}4$), two 32-seed replicates & 64 & 80k & 31/64 & fit/eval \\
Forecast & Ant ($r{=}4$) & 32 & 80k & 11/32 & fit/eval \\
Live deployment & \raggedright Walker2d ($r{=}4$), fresh seeds; 21 run both with and without the controller & 28 & 80k & \textendash & eval \\
Controls & HalfCheetah, Hopper, Humanoid ($r{=}4$) & 12, 16, 12 & 80k & 0/40 & char. \\
Controls & HalfCheetah, Hopper, Humanoid ($r{=}8$) & 6, 4, 4 & 60k & 0/14 & char. \\
Long horizon & Ant ($r{=}4$), forecast seeds 0--7 extended & 8 & 200k & 4/8 & char. \\
Long horizon & HalfCheetah, Humanoid ($r{=}16$) & 4 each & 200k & none & char. \\
Resets & HalfCheetah ($r{=}16$): none, dormancy & 7, 6 & 60k & 0/13 & char. \\
\iflnsection
LayerNorm critic & \raggedright Walker2d ($r{=}4$), LayerNorm vs plain, same seeds (Sec.~\ref{sec:ln}) & 32 pairs & 80k & 0/32, 17/32 & eval \\
\fi
\bottomrule
\end{tabular}
\end{table}

\section{Results}\label{sec:results}
\subsection{Collapse is survivable; fast \texorpdfstring{$\Q$}{|Q|} growth marks the flagged runs}\label{sec:decouple}
\begin{figure}[t]\centering\includegraphics[width=\linewidth]{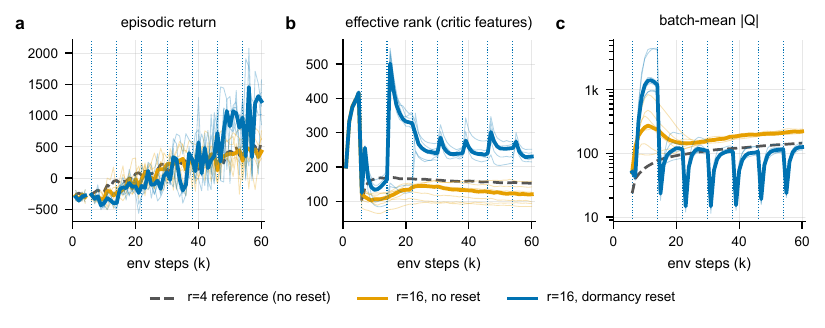}
\caption{\textbf{A collapsed critic with bounded $\Q$ keeps learning.} HalfCheetah at $r{=}16$ to 60k without
resets (7 seeds) and with a dormancy-triggered critic reset (6 seeds; dotted lines mark the resets), and a
reference at $r{=}4$ (12 seeds). Thin lines: seeds; thick: median. (a) Return, (b) effective rank of the critic's
penultimate layer, (c) $\Q$. The guard flags no run.}\label{fig:reset}
\end{figure}

\textbf{Collapse occurs without divergence} (Figures~\ref{fig:overview}A--B and~\ref{fig:reset}). HalfCheetah's critic
collapses at all four UTD ratios, with more than three quarters of its units dormant, yet the guard flags none of
its runs. At $r=16$ its return still rises through 60k while $\Q$ stays of order $10^2$ (Figure~\ref{fig:reset}). A
dormancy-triggered critic reset fires at every opportunity and so acts as a periodic reset
\citep{nikishin2022primacy,doro2023replay}. It lifts the return further, and the reset arm ends ahead on all 6 paired
seeds (Appendix~\ref{app:reset}).

\textbf{$\Q$ growth precedes the flag.} In all 24 \emph{event-timing runs}, flagged runs at $r\ge8$ logged
every 500 steps, $\Q$ departs from its own baseline, taken early in the 10k steps before the flag, a
median 6.31k steps before the flag (at most 7.5k by construction; Appendix~\ref{app:mech}). In the split configurations every run grows $\Q$ (Appendix~\ref{app:labels}). Flagged runs grow it
faster, and the difference shows by 15k.

\textbf{Dormancy carries no warning; the post-learning rank drop co-emerges with $\Q$.} Within the split
configurations, the dormant fraction at 15k forecasts the flag no better than chance (Table~\ref{tab:forecast}b). The
post-learning rank drop at 15k forecasts the flag about as well as $g$ (Table~\ref{tab:forecast}b). In the 24
event-timing runs, rank departs from its own baseline in 14, a median 0.5k steps after $\Q$ (Appendix~\ref{app:mech}).

\textbf{Takeaway:} fast $\Q$ growth marks the flagged runs, and the post-learning rank drop co-emerges with it.
Dormant units, the signal plasticity-loss diagnostics watch, forecast the flag at chance, and the rank drop at
learning onset also appears in HalfCheetah, which the guard never flags.

\subsection{Early \texorpdfstring{$\Q$}{|Q|} growth ranks runs by how soon the guard flags them}\label{sec:forecast}
\begin{figure}[t]\centering\includegraphics[width=\linewidth]{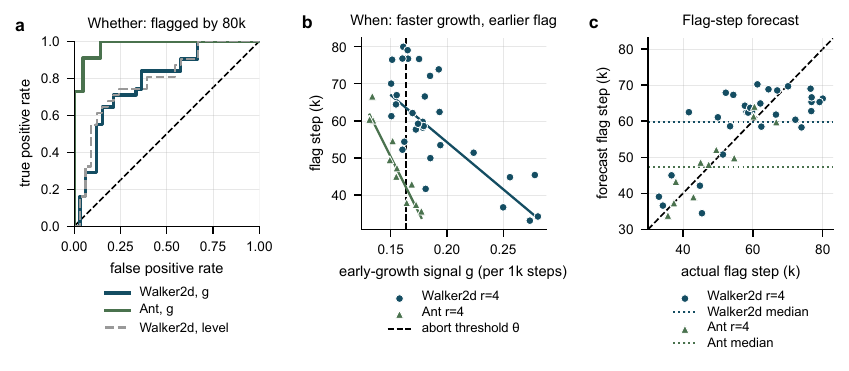}
\caption{\textbf{Faster early growth, earlier flag.} Split configurations, leave-one-seed-out; Table~\ref{tab:forecast}
gives the numbers. (a) ROC of $g$ (both tasks) and of the level $\log_{10}\Q$ at 15k (Walker2d) for a flag by 80k,
one cut of the order that Table~\ref{tab:forecast}a scores. (b) Flag step against $g$
among flagged runs, with Theil--Sen fits; dashed: the live abort threshold $\theta$ (Section~\ref{sec:control}).
(c) Flag step forecast by a LOSO linear regression on $g$; dotted: each task's median flag step, the constant
forecast.}\label{fig:forecast}
\end{figure}

\begin{table}[t]\centering\footnotesize
\caption{\textbf{Early $\Q$ growth ranks runs by how soon the guard flags them.} (a) Harrell's C of $g$ against the
flag step (unflagged runs censored at 80k), over all pairs and over the pairs whose earlier run flags by 40k or by
50k. (b) LOSO AUC for ``flagged by 80k'' at 15k, one cut of that order. (c) Mean absolute error of the LOSO
flag-step regression on $g$ and of a constant median forecast. Brackets: 95\% bootstrap CI over seeds.
Appendix~\ref{app:forecast} defines the $\Q$-free set.}
\label{tab:forecast}
\setlength{\tabcolsep}{4pt}
\begin{tabular}{@{}lcc@{}}
\toprule
 & Walker2d $r{=}4$ (31/64) & Ant $r{=}4$ (11/32) \\
\midrule
\multicolumn{3}{@{}l}{\emph{(a) Time to the flag: Harrell's C of $g$}} \\
All pairs & 0.783 [0.690, 0.864] & 0.983 [0.958, 1.000] \\
Earlier run flags by 40k / by 50k & 0.962 / 0.933 & 0.989 / 0.985 \\
\multicolumn{3}{@{}l}{\emph{(b) Flagged by 80k: LOSO AUC}} \\
Early-growth signal $g$ & 0.780 [0.655, 0.885] & 0.978 [0.926, 1.000] \\
Level $\log_{10}\Q$ at 15k & 0.788 & 0.974 \\
Post-learning rank drop & 0.792 & 0.935 \\
$\Q$-free set of eight statistics & 0.714 & 0.952 \\
Dormant fraction & 0.381 & 0.372 \\
\multicolumn{3}{@{}l}{\emph{(c) Flag step}} \\
Flag-step error, regression on $g$ / constant & 7.85k / 11.31k & 3.13k / 9.43k \\
\bottomrule
\end{tabular}
\end{table}

\textbf{The early-growth signal ranks the runs by time to the flag, the soonest flags best}
(Table~\ref{tab:forecast}a). We score it as a time to an event: we censor a run the guard has left unflagged
at 80k, and 30 of the 33 such Walker2d runs were still climbing (Appendix~\ref{app:labels}). Harrell's C of $g$ is
0.783 on Walker2d and 0.983 on Ant, and it is highest for the pairs whose earlier run flags soonest: the three
Walker2d runs flagged by 40k rank 2nd, 4th and 6th of 64 by $g$. A threshold rule aborts the top of this ranking.
The AUC for a flag within a budget is one cut of the order (Figure~\ref{fig:forecast}a).
On Walker2d it falls from 0.956 at 40k to 0.780 at 80k, since the late flags start with early growth close to that
of the runs still unflagged at 80k. The order holds on every plain-critic SAC cohort of the split configurations
collected after we fixed the window: the Ant cohort and, on Walker2d, the second replicate, the live
seeds\iflnsection{} and the plain arm of Section~\ref{sec:ln}\fi{} (Appendix~\ref{app:horizon}), although on
Walker2d these score lower than the first replicate. It also holds for other early windows and a lower flag
threshold (Appendix~\ref{app:forecast}).

\textbf{The forecast is no extrapolation of $\Q$ to a threshold.} Reaching the level is no guarantee of a flag: 18 of
the 33 unflagged Walker2d runs end the budget at or above the lowest flag level. The trend misleads too: the early climb, extended
as a straight line, reaches the flag level a median 32k steps too soon on Walker2d and loses to a constant forecast
(Section~\ref{sec:mechanism}), while the order of the climbs ranks the flags. From 15k, a flagged run's
$\Q$ still grows a median 122-fold on Walker2d (20- to 255-fold, 18.1k to 65.0k steps later) and 389-fold on Ant
(164- to 570-fold). At 15k the largest $\Q$ of any run lies 7.9-fold (Walker2d) and 145-fold (Ant) below the lowest
$\Q$ at which a run of the cohort flags. Seeds with identical settings split, and the first 15k of 80k steps already
order them; dormancy, the diagnostic plasticity-loss work watches, forecasts at chance there
(Section~\ref{sec:decouple}). The guard's loss criterion fires first in 155 of the 157 flags
(Appendix~\ref{app:labels}).

\textbf{One number read at 15k carries the forecast.} The level $\log_{10}\Q$ at 15k forecasts as well as $g$
(Table~\ref{tab:forecast}b): the starts are close (Section~\ref{sec:mechanism}), so the level is mostly accumulated
rate, and in the data the two are nearly collinear, although in sample the level adds information to $g$
(Appendix~\ref{app:nested}). Given the level, the fitted coefficient of $g$ is negative, the opposite of the rate
model's prediction, so the pair encodes one quantity, how far the climb has come. The controller uses $g$, fixed in advance.

\textbf{It also forecasts the flag step} (Table~\ref{tab:forecast}c). Among flagged runs, faster early growth goes
with an earlier flag (Figure~\ref{fig:forecast}b; Section~\ref{sec:mechanism} gives the rank correlations), and a
LOSO regression of the flag step on $g$ beats a constant median forecast on both tasks (Figure~\ref{fig:forecast}c).

\textbf{Takeaway:} long before the flag, early $\Q$ ranks the seeds of a split configuration by how soon the guard
flags them, on Ant almost without error. Section~\ref{sec:mechanism} models this order;
Section~\ref{sec:control} acts on it.

\subsection{A shared-profile model is consistent with the direction of the flag order}\label{sec:mechanism}
\textbf{A rate model: each run climbs at its own rate on a shared profile.} Linearized analyses of temporal-difference learning tie divergence to an
update that multiplies the values by a constant ratio at every step \citep{achiam2019characterizing}, which would
make $\log_{10}\Q$ a straight line in time. For multilayer critics trained offline with Adam, SEEM instead predicts
growth polynomial in the number of updates, which bends in $\log\Q$ \citep{yue2023seem}. Our runs climb and then
bend, so each run gets its own rate on a profile $h$, shared within a configuration, that absorbs the
bend. Run $i$ follows
\begin{equation}
\log_{10}\Q_i(t)=a_i+g_i\,h(t),\qquad T_{{\rm flag},i}=h^{-1}\big((\log_{10}\tau-a_i)/g_i\big).
\label{eq:rate}
\end{equation}
Here $a_i$ and $g_i$ are the run's start and rate, and $\tau$ is the common $\Q$ at which the guard flags a run. The
profile $h$ is any increasing curve, scaled so that $g_i$ is the early-growth signal. In the data the flagged runs do flag near
one level, of order $10^5$: the loss criterion fires while $\Q$ is in that band (Section~\ref{sec:setup}). Reaching the band is no guarantee of a flag
(Section~\ref{sec:forecast}). Appendix~\ref{app:rate} proves two consequences for the flagged runs. From a common start, a faster early rate brings an earlier flag, whatever the shape of $h$.
If $h$ bends downward, the early straight line reaches $\tau$ before the flag.

\textbf{Faster early growth, earlier flag; the order is loose on Walker2d.} In the split configurations the starts
spread under a third as widely as the early climbs, so the early level and $g$ rank the runs alike. On Walker2d the $\Q$ in the telemetry record at the flag stays
within threefold across the flagged runs (Appendix~\ref{app:rate}). In the two split configurations and Walker2d at
$r=8$, whose flags all come after the early window, the Spearman correlation between $g$ and the flag step lies between
$-0.64$ and $-0.96$, every 95\% interval below zero, so the model gets the direction of the order right. On Walker2d
at $r=4$ the order is looser than the model implies: from the measured starts and a common flag level, the model's
order correlates with $g$ at $-0.987$ against the observed $-0.64$, and these starts and this level leave the
out-of-order pairs unexplained (Appendix~\ref{app:rate}).

\textbf{The climb bends away from a straight line.} A straight line in $\log_{10}\Q$ (a linear $h$, constant-ratio
growth) fits the climb to the flag well by $R^2$, with medians of 0.95--0.99 in the tested cohorts outside Walker2d at
$r=4$ and 0.891 there (Appendix~\ref{app:rate}). But the climb bends: a quadratic fit to $\log_{10}\Q$ before the
flag curves downward in all 72
flagged runs of the split configurations and the phase map. The straight line extended from $t_d$ therefore lands
early, by a median 32k steps on Walker2d and 16k on Ant, and forecasts the flag step worse than the constant forecast,
which the regression on $g$ beats (Table~\ref{tab:forecast}c).

\textbf{The post-learning rank drop moves with the rate; dormancy does not.} At $t_d$, the post-learning rank drop rises with
$g$ (Spearman $+0.59$ on Walker2d and $+0.95$ on Ant), while the dormant fraction shows no such relation
(Appendix~\ref{app:rate}).\iflnsection{} A LayerNorm critic lowers the rate itself (Section~\ref{sec:ln}).\fi

\textbf{Takeaway:} runs start alike and the guard fires near a common $\Q$, so faster climbs tend to flag first,
loosely on Walker2d. The climb bends: a flag step takes a fitted map, an abort only the order.

\subsection{The growth-abort controller saves compute out of sample and live}\label{sec:control}
\textbf{The controller aborts on $g$ at 15k.} At $t_d$, the growth-abort controller computes $g$ from
the run's own log and aborts the run if $g$ exceeds the abort threshold $\theta$. We fit the threshold to maximize net
steps saved on training seeds (Algorithm~\ref{alg:ctrl}, Appendix~\ref{app:ctrl}). Our compute ledger follows the
standard evaluation protocol, which scores a run by its policy at the end of training \citep{agarwal2021precipice}; the
guard stops a flagged run before it
gets there, so the ledger counts that run as lost (Appendix~\ref{app:alt} values its best checkpoint instead). The
controller's choice barely depends on how well a run learns: over the 64 Walker2d runs, $g$ and the best logged return are nearly
uncorrelated (Spearman $+0.14$, $p=0.25$). A true abort saves the $T_{\rm flag}-t_d$
steps the run would have spent before its flag, and each false abort costs the full budget $H=80$k. The oracle aborts
every flagged run and no other.

\begin{table}[h]\centering\footnotesize
\caption{\textbf{The growth-abort controller saves compute out of sample and live.} Net environment steps saved
under the compute ledger of Section~\ref{sec:control} (decision at 15k, $H=80$k). TP/FP/FN: true aborts, false
aborts, missed flags; Oracle: aborts every flagged run and no other. (a) 8-fold rows are out of sample; we fixed the
live abort threshold in advance on the 64 Walker2d seeds. (b) Net steps saved, in
thousands, by abort rules on the same folds; the two label-free rules fit no threshold. Appendix~\ref{app:ctrl} gives
every ledger row and the TP/FP/FN counts of (b).}
\label{tab:ctrl}
\setlength{\tabcolsep}{3.5pt}
\begin{tabular}{@{}lrcccrr@{}}
\toprule
(a) Cohort & $n$ & TP/FP/FN & Prec. & Rec. & Net saved & Oracle \\
\midrule
SAC Walker2d $r{=}4$, 8-fold & 64 & 22/6/9 & 0.79 & 0.71 & 444{,}109 & 1{,}391{,}847 \\
SAC Ant $r{=}4$, 8-fold & 32 & 10/1/1 & 0.91 & 0.91 & 247{,}386 & 372{,}819 \\
Live, 21 paired fresh Walker2d seeds & 21 & \textendash & \textendash & \textendash & 130{,}085 & 512{,}263 \\
Live abort threshold, all 28 fresh seeds & 28 & \textendash & \textendash & \textendash & 252{,}389 & 803{,}515 \\
\bottomrule
\end{tabular}

\vspace{4pt}
\begin{tabular}{@{}lcc@{}}
\toprule
(b) Abort rule, read at 15k & Walker2d $r{=}4$ & Ant $r{=}4$ \\
\midrule
Early-growth signal $g$ (the controller) & 444k & 247k \\
Level $\log_{10}\Q$ & 333k & 247k \\
Post-learning rank drop & 438k & 151k \\
Dormant fraction & $-$261k & $-$80k \\
Low return & 103k & $-$80k \\
Median stopping rule on return, label-free & $-$588k & $-$706k \\
$g$ above the cohort's 75th percentile, label-free & 151k & 231k \\
\bottomrule
\end{tabular}
\end{table}

\textbf{It saves compute on held-out seeds} (Table~\ref{tab:ctrl}a, Figure~\ref{fig:decide}). Under 8-fold
cross-validation, the controller saves 9.9\% of the Walker2d cohort's compute (its 64 runs, each to its
flag or to 80k) and 31.9\% of what the oracle saves. On Ant it saves 66.4\% of the oracle's saving. The Walker2d saving
holds across random fold splits (Appendix~\ref{app:ctrl}). A label-free variant aborts a run whose
$g$ lies above the 75th percentile of the other runs' $g$. In both split configurations it also saves compute
(Table~\ref{tab:ctrl}b).

\textbf{It saves more than rules that watch dormancy or return} (Table~\ref{tab:ctrl}b). On the same folds and
ledger, the dormant-fraction rule and the median stopping rule on return \citep{golovin2017vizier} lose compute on
both tasks. The low-return rule saves far less than $g$ on Walker2d and loses on Ant. The post-learning rank drop, which
co-emerges with $\Q$ growth (Section~\ref{sec:decouple}), saves nearly as much as $g$ on Walker2d and less on Ant,
and it needs a probe batch and an SVD.

\textbf{It saves compute live on fresh seeds.} We analyze 21 fresh Walker2d seeds, each run without and
with the controller, $\theta$ fixed in advance. On one machine, runs are deterministic given the seed
(Appendix~\ref{app:live}), so each uncontrolled twin shows the fate of its controlled run. On these 21 pairs the
controller saves 25.4\% of the oracle's saving, and $\theta$ also saves compute on all 28 uncontrolled fresh runs
(Table~\ref{tab:ctrl}a).

\textbf{Takeaway:} calibrated once on labeled runs, the abort threshold on $g$ turns the forecast
into saved compute on held-out and fresh seeds. It beats every abort rule that watches dormancy or return.

\iflnsection
\subsection{A LayerNorm critic removes the flag on Walker2d at \texorpdfstring{$r=4$}{r=4}}\label{sec:ln}
We repeat the Walker2d $r=4$
study with layer normalization after each critic hidden layer, as in SimBa, BRO and RLPD
\citep{lee2025simba,nauman2024bro,ball2023rlpd}, keeping the network, label, budget ($H=80$k) and window. Each of 32
fresh seeds runs twice on one GPU, once with each critic, and an exact McNemar test compares the paired flags.
The guard flags none of the 32 LayerNorm runs and 17 of the 32 plain runs,
so every discordant pair is a flagged plain run.

\begin{figure}[h]\centering\includegraphics[width=\linewidth]{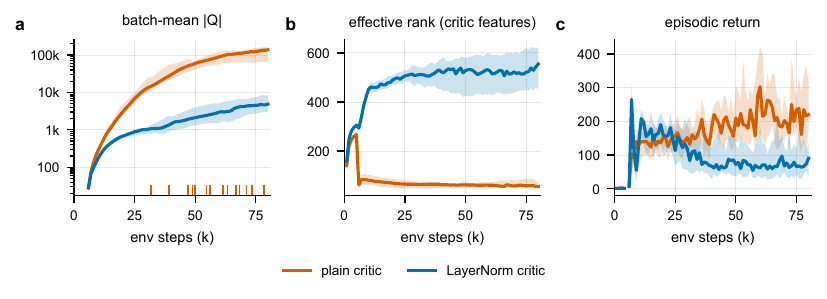}
\caption{\textbf{LayerNorm keeps $\Q$ lower and the critic's rank high.} Walker2d $r{=}4$, the 32 seed pairs of
Table~\ref{tab:ln}: median (line) and interquartile range (band) over the runs still training. The guard
stops a flagged plain run at its flag (ticks in a), so late plain-critic curves describe unflagged runs. (a) Batch-mean
$\Q$, (b) effective rank of the critic's penultimate layer, (c) return.}\label{fig:ln}
\end{figure}

\textbf{LayerNorm lowers the rate $g$ and keeps the critic's rank high} (Figure~\ref{fig:ln}). It lowers $g$ in 31
of the 32 pairs (median 0.121 against 0.160), and the LayerNorm critic ends with a median dormant fraction near zero
(Appendix~\ref{app:ln} gives the medians). LayerNorm thus lowers $\Q$ growth, the post-learning rank drop and dormancy
together. HalfCheetah (Section~\ref{sec:decouple}) separates $\Q$ growth from dormancy and the rank drop at learning
onset; the post-learning drop co-emerges with it.

\textbf{LayerNorm lowers the return} (Figure~\ref{fig:ln}c). On the pairs in which the guard flags neither arm, the
LayerNorm critic ends at about a third of the plain critic's return (Table~\ref{tab:ln}).

\begin{table}[h]\centering\footnotesize
\caption{\textbf{A LayerNorm critic removes the flag at $r{=}4$.} Walker2d $r{=}4$, all 32 seed pairs. Exact
McNemar test on the discordant pairs (the guard flags the plain run in all 17; $p=1.5\times10^{-5}$), with Wilson
intervals. The last row is the plain-critic cohort of Section~\ref{sec:forecast}, the reference for the plain arm
(Fisher $p=0.83$). Return: on the 15 pairs flagged in neither arm, the mean return over 70k--80k has median 84.8
with LayerNorm against 251.7 without (Wilcoxon signed-rank $p=0.0015$).}
\label{tab:ln}
\setlength{\tabcolsep}{4pt}
\begin{tabular}{@{}llcc@{}}
\toprule
Critic & Seeds & $H$ & Flagged by $H$ (Wilson 95\%) \\
\midrule
LayerNorm & 32 pairs & 80k & 0/32 [0.000, 0.107] \\
Plain (same seeds) & 32 pairs & 80k & 17/32 [0.364, 0.691] \\
Plain, the 64 seeds of Section~\ref{sec:forecast} & 64 & 80k & 31/64 [0.366, 0.604] \\
\bottomrule
\end{tabular}
\end{table}

\textbf{Takeaway:} on Walker2d at $r=4$, a LayerNorm critic removes the flag at the cost of return. For the plain
critic, the growth-abort controller cuts the cost of the flag instead (Section~\ref{sec:control}).
\fi

\section{Scope and Outlook}\label{sec:limits}
\textbf{Scope.} Our evidence spans five MuJoCo tasks, four UTD ratios and 612 runs of scaled critics (SAC and TD3),
with\iflnsection{} a LayerNorm critic and\fi{} a standard-width reference as further tests. The practical saving rests on Walker2d,
since no Ant run at $r=4$ reached a positive return; the forecast is sharpest on Ant all the same, so it tracks the
critic itself.
With standard $256\times2$ networks and batch 256, the same trainer flags none of 24 Walker2d runs at $r=16$ in 40k
steps, against all 10 phase-map runs of the $2048\times2$ networks, whose median $g$ is about four times as large
(Appendix~\ref{app:stdwidth}). The flag thus comes with scaling up the networks and batch, and high-UTD methods scale
critic size \citep{nauman2024bro,lee2025simba}.
We test the order of the flags in the two split configurations and in
Walker2d at $r=8$ (Section~\ref{sec:mechanism}), and compare rates within a configuration, since the rate rises with $r$ in Walker2d and Ant and falls in Humanoid
(Appendix~\ref{app:rate}).

\textbf{Outlook.} Longer horizons come next. HalfCheetah shows collapse to be survivable through 60k, and a
HalfCheetah run at $r=16$ continued to 200k grows $\Q$ past $10^3$ and ends below half its best return without a flag
(Appendix~\ref{app:labels}). For TD3, the guard flags Walker2d runs only at $r=16$, and there the forecast is
indistinguishable from chance and the controller loses compute (Appendix~\ref{app:td3}). CrossQ, BRO, SimBa,
MAD-TD\iflnsection{}, LayerNorm above $r=4$\else{}, LayerNorm\fi{} and discrete actions, where a MinAtar DQN pilot
never split (Appendix~\ref{app:dqn}), are the next tests.

\section{Conclusion}
Two symptoms grouped under ``plasticity loss'' behave differently at high UTD ratios: a HalfCheetah critic with
dormant units and a rank drop at learning onset keeps learning while its $\Q$ stays bounded. Within one
high-UTD SAC configuration, runs start close together, and how far a critic's $\log_{10}\Q$ has climbed by step 15k, its early
growth, ranks the runs by how soon the guard flags them. On Walker2d and Ant at $r=4$ this rate orders the flags out of sample, and the growth-abort
controller turns the forecast into saved compute on held-out and fresh Walker2d seeds.\iflnsection{} On Walker2d at
$r=4$, a LayerNorm critic lowers the rate and removes the flag at a cost in return.\fi{}

\clearpage
\label{lastmainpage}%
\label{firstpageafterbody}

\subsection*{AI use statement}

Generative AI tools were used solely to assist with language editing and improve the clarity and
readability of the manuscript. All scientific content, analyses, results, interpretations, and
conclusions were developed and verified by the authors, who take full responsibility for the
content of the paper.

\subsection*{Ethics statement}

This work studies the training stability of off-policy reinforcement-learning agents on simulated
MuJoCo locomotion tasks from public benchmark suites. It involves no human subjects, no personal or
sensitive data, and no deployment outside simulation. Its practical output is a monitoring and
early-stopping rule whose direct effect is to spend less compute on training runs that end in value divergence.
The foreseeable risk is over-trust: the controller aborts some runs the guard would not have flagged by 80k and
misses some it would. Its
benefit holds only in the configurations where we calibrated it (Section~\ref{sec:control}), and it offers no
general guarantee of training stability.

\subsection*{Reproducibility statement}

Appendix~\ref{app:impl} lists every hyperparameter (identical across tasks, with no per-task tuning), the
probes and the exact divergence label. Appendix~\ref{app:labels} gives the label census and the zero-event
bounds; Appendix~\ref{app:forecast} the early-growth signal, the leave-one-seed-out evaluation, the window,
horizon and nested-model analyses and the calibration measure; Appendix~\ref{app:ctrl} the growth-abort controller as
pseudocode, the accounting ledger, the live-deployment protocol and the transfer test; and
Appendix~\ref{app:compute} the hardware, software, seeding and which runs replay others. The training code, the analysis and
figure-building scripts, the per-seed run logs (configuration, outcome, and the logged critic and return
telemetry) and the text output of every analysis are available from the corresponding author on request; with
them, every reported number can be recomputed from the logs.

\bibliographystyle{iclr2027_conference}
\label{referencesstart}%
\bibliography{refs}

\newpage
\appendix

\setcounter{figure}{0}
\renewcommand{\thefigure}{S\arabic{figure}}
\renewcommand{\theHfigure}{S\arabic{figure}}
\setcounter{table}{0}
\renewcommand{\thetable}{S\arabic{table}}
\renewcommand{\theHtable}{S\arabic{table}}

\section{Implementation and Hyperparameters}\label{app:impl}
All runs use one implementation with the settings in Table~\ref{tab:hyper}. SAC and TD3 share the
optimizer, discount, target-smoothing rate, replay buffer, warm-up and the scaled networks (two hidden
layers of width 2048, no normalization layer, batch 2048); the UTD ratio $r$ is the swept axis. SAC uses twin
critics with clipped double-$Q$ targets, a tanh-Gaussian actor and automatic entropy tuning (target entropy
$-\dim\mathcal{A}$). TD3 uses a deterministic actor with target-policy smoothing, clipped double-$Q$ and
delayed policy updates.

\paragraph{Telemetry.} The trainer writes one record every 1k environment steps, with two exceptions. It writes
one every 500 steps in the event-timing runs, the clipping runs and the phase-map runs of Walker2d at $r\ge4$, Ant
at $r\le8$, and HalfCheetah, Hopper and Humanoid at $r=16$; and one every 2k in the second Walker2d replicate, the
80k Ant cohort at $r=4$ and the runs to 200k. A record holds
the mean episodic return in the window, the mean critic loss (the sum of both critics' mean squared TD errors), and
$\Q$, the mean over the window's gradient steps of the batch-mean $|Q_1(s,a)|$. At the end of each window
the trainer computes the effective rank \citep{roy2007effrank} and the dormant fraction \citep{sokar2023dormant}
(thresholds $0$ and $0.1$) of the first critic's penultimate activations for a held-out probe batch of 2048
state--action pairs.

\paragraph{Divergence guard.} After every gradient step the trainer checks the critic loss and the
batch-mean $|Q_1|$. The guard flags and stops a run at the first step at which either is non-finite, the
loss exceeds $10^8$, or the batch-mean $|Q_1|$ exceeds $10^6$, and the trainer records that step as $T_{\rm flag}$.

\paragraph{Resets (Figure~\ref{fig:reset}).} The reset re-initializes both critics and their target
networks and clears the critic optimizer state. The dormancy trigger fires when the fraction of 0.1-dormant
units exceeds 0.1. A second arm used a rank trigger (effective rank below 0.6 of its running peak over
post-learning probes), with at least 8k steps between resets. The rank trigger never fired: the rank falls as
learning starts (median 414.8 before, 120.3 at the first post-learning probe), the running peak comes after
that fall, and later ranks stay at 0.73 or more of it. That arm therefore matches the no-reset arm, and we omit
it.

\begin{table}[h]\centering\small
\caption{Hyperparameters, identical across tasks (no per-task tuning). Unlisted values follow the
standard SAC and TD3 recipes \citep{haarnoja2018sac,fujimoto2018td3}.}
\label{tab:hyper}
\begin{tabular}{@{}lll@{}}
\toprule
Hyperparameter & SAC & TD3 \\
\midrule
Optimizer, learning rate & Adam, $3\times10^{-4}$ & Adam, $3\times10^{-4}$ \\
Discount $\gamma$; target smoothing & $0.99$; $0.005$ & $0.99$; $0.005$ \\
Replay buffer; warm-up steps & $10^6$; 5000 & $10^6$; 5000 \\
Batch size & 2048 & 2048 \\
Critic and actor width $\times$ depth & $2048\times2$, no normalization & $2048\times2$, no normalization \\
UTD ratio $r$ & $\{2,4,8,16\}$ & $\{4,8,16\}$ \\
Critics & twin (clipped) & twin (clipped) \\
Entropy & auto, target $-\dim\mathcal{A}$ & none \\
Target policy noise & none & $0.2$ (clip $0.5$) \\
Policy delay; exploration noise & none & 2; $0.1$ \\
Probe batch & 2048 & 2048 \\
Divergence guard & loss $>10^8$, $|Q_1|>10^6$ or non-finite & same \\
\bottomrule
\end{tabular}
\end{table}

\section{Divergence Label, Censoring, Returns and Zero-Event Bounds}\label{app:labels}
\begin{figure}[h]\centering\includegraphics[width=\linewidth]{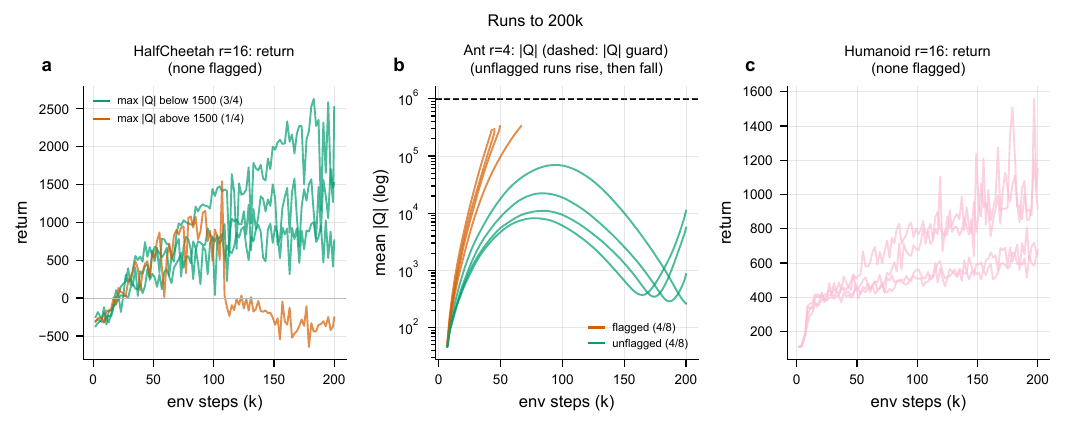}
\caption{Runs to 200k. (a) HalfCheetah $r{=}16$, the 4 runs to 200k (seeds 0, 1, 5 and 6): the guard flags none;
color marks the one whose $\Q$ exceeded $1.5\times10^3$ (seed 6; Table~\ref{tab:hclong}). (b) Ant
$r{=}4$, seeds 0--7 of the forecast cohort continued: 4 flagged, all before 80k; the 4 unflagged runs' $\Q$
rises to a peak between 77k and 95k and then falls more than tenfold (Table~\ref{tab:reversal}). The dashed
$10^6$ line is the $\Q$ guard, which these runs never reached (the loss guard flagged them at $\Q$ of order
$10^5$). (c) Humanoid $r{=}16$, the 4 runs to 200k (seeds 0, 1, 5 and 6): the guard flags none, maximum $\Q$ between 137
and 158.}\label{fig:S-long}
\end{figure}

\paragraph{What fires the guard.} Of the 157 flagged runs, the loss criterion fired first in 155
and none went non-finite. In every flagged run without gradient clipping the loss fired while $\Q$ was of
order $10^5$: the largest batch-mean $\Q$ before the flag has median $2.06\times10^5$ (range
$1.06\times10^5$--$2.82\times10^5$) on Walker2d at $r=4$ and $3.43\times10^5$ ($2.36\times10^5$--$4.67\times10^5$)
on Ant at $r=4$. The $\Q$ criterion fired first in two gradient-clipped Ant runs at clip 1 (batch-mean
$|Q_1|$ of $1.0021\times10^6$ and $1.0001\times10^6$ while the loss was $7.71\times10^7$ and $4.31\times10^7$).
The $10^6$ lines drawn in Figures~\ref{fig:S-long} and~\ref{fig:S-anatomy} mark a criterion that none of those
unclipped runs reached.
At the decision step the flagged runs sit far below this band. From the record at 15k to the flag record, the
window-mean $\Q$ of a flagged run grows a median 121.6-fold on Walker2d at $r=4$ (range 20.3--255.3; 95\% CI of the
median 99.3--151.2), or 2.08 orders of magnitude, over a median 44.76k steps, and 389.3-fold on Ant at $r=4$
(163.6--570.4; CI 249.3--484.1), or 2.59 orders of magnitude, over a median 32.34k steps. At 15k the largest $\Q$
of any run in the cohort, flagged or not, is $1.28\times10^4$ on Walker2d and $1.57\times10^3$ on Ant: 7.9-fold and
144.7-fold below the lowest $\Q$ in a flag record ($1.01\times10^5$ and $2.27\times10^5$).

\paragraph{A lower threshold.} Counting a run as positive when the guard flags it or when its window-mean $\Q$
after 15k reaches $10^5$ gives 49 of 64 positives on Walker2d and 14 of 32 on Ant (50 and 15 if any logged
per-step maximum counts). The Walker2d LOSO AUC of $g$ under this label is 0.848. At $10^4$, all 64 Walker2d runs
and 29 of the 32 Ant runs are positive, and no AUC exists.

\paragraph{Censoring and the runs followed to 200k.} Every unflagged run is right-censored at its horizon. The
33 unflagged Walker2d runs at $r=4$ end at a median $\Q$ of $1.04\times10^5$ (maximum $2.17\times10^5$), and 30 of
them were still rising over their last 20k steps. All 21 unflagged Ant runs were still rising (median final
$\Q$ $3.72\times10^4$). The divergence-free tasks end at $\Q$ between 132 and $1.01\times10^3$ (median over runs
per cell). Ant at $r=4$ to 200k (seeds 0--7 of the forecast cohort, continued) flagged 4 of 8 runs, all before
80k (at 42,969 to 66,688 steps), and none of the other 4 later (Wilson upper bound 0.490).
Table~\ref{tab:reversal} shows what those 4 did instead: $\Q$ rose to a first peak of $8.18\times10^3$ to
$6.97\times10^4$ between 77k and 95k, then fell more than tenfold, to 264--371, and in 3 of the 4 it was rising
again at 200k (Figure~\ref{fig:S-long}b). Of all unclipped SAC run records, 246 passed $\Q=10^4$, and only
three of them (Ant seeds 2, 3 and 4 in this cohort) later fell tenfold below an earlier peak. The guard stops every
run it flags, so we observe no run after its flag.

\paragraph{HalfCheetah to 200k.} These runs bound in time the survivability claim of Section~\ref{sec:decouple}.
The guard flags none of the 4 HalfCheetah runs at $r=16$ run to 200k (seeds 0, 1, 5 and 6;
Table~\ref{tab:hclong}). Seed 6 ends below half its best 20k-window median return. It is the run whose
$\Q$ grew past $10^3$ (maximum $6.24\times10^3$), and it ends at the lowest return, $-354.5$; the other three kept
$\Q$ at most 306. The guard also flags none of the 4 Humanoid runs at $r=16$ run to 200k (seeds 0, 1, 5 and 6), whose
maximum $\Q$ lies between 137 and 158.

\begin{table}[h]\centering\footnotesize
\caption{Ant $r{=}4$ runs continued to 200k: window-mean $\Q$ at the first peak (the maximum up to 120k), the
lowest value after it, and the last value.}
\label{tab:reversal}
\begin{tabular}{@{}lccccc@{}}
\toprule
Seed & Flagged & First peak (step) & Lowest after it (step) & At 200k \\
\midrule
0, 1, 5, 6 & at 45,030; 49,509; 66,688; 42,969 & & & \\
2 & no & $1.1\times10^4$ (85k) & 346 (175k) & $5.58\times10^3$ \\
3 & no & $6.97\times10^4$ (95k) & 264 (200k) & 264 \\
4 & no & $2.22\times10^4$ (85k) & 287 (187k) & 855 \\
7 & no & $8.18\times10^3$ (77k) & 371 (165k) & $1.11\times10^4$ \\
\bottomrule
\end{tabular}
\end{table}

\begin{table}[h]\centering\footnotesize
\caption{HalfCheetah $r{=}16$ runs to 200k: maximum window-mean $\Q$, median return over the last 20k steps, and
best 20k-window median return, the last window included. The guard flags none.}
\label{tab:hclong}
\setlength{\tabcolsep}{5pt}
\begin{tabular}{@{}lcccc@{}}
\toprule
Seed & 0 & 1 & 5 & 6 \\
\midrule
Max $\Q$ & 266 & 306 & 245 & $6.24{\times}10^3$ \\
Last 20k return & 2320.8 & 805.6 & 1459.5 & $-354.5$ \\
Best 20k return & 2320.8 & 884.9 & 1459.5 & 931.6 \\
\bottomrule
\end{tabular}
\end{table}

\paragraph{Returns.} Table~\ref{tab:returns} gives the returns per task. Returns are low at these budgets for
every task (median best return at $r=4$ within 80k: 449 on Walker2d, 948 on HalfCheetah, 662 on Hopper, 529 on
Humanoid). Walker2d runs at $r=4$ learn: the median
best logged return is 423.6 for unflagged and 485.9 for flagged runs, and the median final return (each run's median
over its last five records) 209.5 and 227.9; Table~\ref{tab:returns} gives the last record instead. No Ant run at $r=4$ ever logged a positive return within 80k (median final return $-167.1$ unflagged,
$-168.0$ flagged). These returns are those of an unnormalized wide critic over short budgets, and they do not measure
tuned SAC; the standard-width reference below compares flags and $\Q$ only.

\paragraph{A standard-width reference.}\label{app:stdwidth} We trained 24 Walker2d runs at $r=16$ for 40k steps with
the standard $256\times2$ actor and critics and batch 256, and with the trainer, the other hyperparameters and the
guard of the main runs. The guard flags none of them (0/24, Wilson 95\% interval [0.000, 0.138]). The largest $\Q$ a
run reaches has median 349 across runs (maximum $2.29\times10^3$), and the median $\Q$ is 264 at 15k, 268 at 30k and
213 at 40k. Over the $r=16$ window $(5\text{k},7.5\text{k}]$ the median early growth $g$ is $2.54\times10^{-4}$ in
$\log_{10}\Q$ per step, against $9.62\times10^{-4}$ for the $2048\times2$ networks with batch 2048 at the same ratio, where the guard
flags all 10 phase-map runs within 30k (Table~\ref{tab:whenr} gives the flag timing of that cell). The comparison
changes actor width, critic width and batch at once; it isolates none of the three. With no flagged run, there is
no forecast to test. The check covers one task, one UTD ratio and 40k steps; standard-width SAC may still diverge on
other tasks, at other ratios or over longer budgets.

\begin{table}[h]\centering\footnotesize
\caption{Median over runs of the best and of the last logged return (window mean).}
\label{tab:returns}
\begin{tabular}{@{}llccc@{}}
\toprule
UTD ratio, horizon & Task & Runs (flagged) & Best & Last \\
\midrule
$r{=}4$, 80k & Walker2d & 64 (31) & 448.9 & 202.0 \\
 & Ant & 32 (11) & $-36.5$ & $-172.0$ \\
 & HalfCheetah & 12 (0) & 947.9 & 747.8 \\
 & Hopper & 16 (0) & 662.1 & 475.6 \\
 & Humanoid & 12 (0) & 528.8 & 453.0 \\
$r{=}8$, 60k & HalfCheetah & 6 (0) & 621.2 & 138.1 \\
 & Hopper & 4 (0) & 590.4 & 323.8 \\
 & Humanoid & 4 (0) & 537.6 & 429.8 \\
\bottomrule
\end{tabular}
\end{table}

\paragraph{Zero-event cells.} Table~\ref{tab:wilson} gives Wilson 95\% intervals. A cell with no flag
bounds the flag rate below 0.10 only with at least 35 runs; every single zero-event cell we ran is smaller.

\begin{table}[h]\centering\small
\caption{Flag rates with Wilson 95\% intervals.}
\label{tab:wilson}
\begin{tabular}{@{}lccc@{}}
\toprule
Cell ($H$) & Flagged & Rate & Wilson 95\% \\
\midrule
SAC Walker2d $r{=}4$, both replicates (80k) & 31/64 & 0.484 & [0.366, 0.604] \\
SAC Ant $r{=}4$ (80k) & 11/32 & 0.344 & [0.204, 0.517] \\
SAC HalfCheetah, Hopper, Humanoid $r{=}4$ (80k) & 0/12, 0/16, 0/12 & 0 & $\le$ 0.243, 0.194, 0.243 \\
\quad the three pooled & 0/40 & 0 & $\le$ 0.088 \\
SAC HalfCheetah, Hopper, Humanoid $r{=}8$ (60k) & 0/6, 0/4, 0/4 & 0 & $\le$ 0.390, 0.490, 0.490 \\
\quad the three pooled & 0/14 & 0 & $\le$ 0.215 \\
SAC HalfCheetah $r{=}16$ (30k) & 0/32 & 0 & $\le$ 0.107 \\
SAC HalfCheetah, all four $r$ (30k phase map) & 0/56 & 0 & $\le$ 0.064 \\
SAC Ant $r{=}4$, flags between 80k and 200k & 0/4 & 0 & $\le$ 0.490 \\
SAC Walker2d $r{=}16$, width 256, batch 256 (40k) & 0/24 & 0 & $\le$ 0.138 \\
TD3 Walker2d and Ant, $r{=}4$; $r{=}8$ (80k) & 0/16; 0/16 & 0 & $\le$ 0.194 \\
TD3 Walker2d and Ant, $r{=}4$ and $r{=}8$ pooled (80k) & 0/32 & 0 & $\le$ 0.107 \\
\bottomrule
\end{tabular}
\end{table}

\section{Forecast Details}\label{app:forecast}
\paragraph{Features.} The early-growth signal $g$ is the least-squares slope of $\log_{10}\Q$ against environment
steps over the records with $5\text{k}<\text{step}\le15\text{k}$ (5 to 10 records per run). The early level
is $\log_{10}$ of the last $\Q$ in the same window. The $\Q$-free set holds the slope and the last value of
effective rank, dormant fraction, return and weight norm over the same window (8 features). The guard
flags no run before 33,099 steps (Walker2d) or 35,675 (Ant), so in the split configurations every window ends before
every flag.

\paragraph{Classifier and intervals.} We standardize each feature set on the training folds and feed it to an
L2-regularized logistic regression, fit leave-one-seed-out within one (task, $r$) configuration, and compute AUC
on the pooled held-out scores. For $g$, LOSO pooling lowers the AUC slightly: the AUC of $g$ itself, with
the sign fixed in advance, is 0.807 on Walker2d (LOSO 0.780) and 0.996 on Ant (0.978). For an uninformative
statistic, LOSO pooling pushes the AUC below 0.5 (Appendix~\ref{app:rankfeat}). CIs are percentile bootstraps
over seeds (2000 resamples, stratified by label) of the fixed held-out scores; we do not refit the classifier
inside the bootstrap, so the intervals understate the uncertainty. Paired AUC differences use DeLong's test; we
compare nested models by likelihood-ratio tests, since DeLong's test is unreliable for them
\citep{demler2012misuse}. Table~\ref{tab:forecastci} gives the intervals of Table~\ref{tab:forecast}b.

\begin{table}[h]\centering\footnotesize
\caption{Table~\ref{tab:forecast}b with 95\% bootstrap CIs over seeds: LOSO AUC for ``flagged by 80k'' at 15k.}
\label{tab:forecastci}
\setlength{\tabcolsep}{4pt}
\begin{tabular}{@{}lcc@{}}
\toprule
 & Walker2d $r{=}4$ (31/64) & Ant $r{=}4$ (11/32) \\
\midrule
Early-growth signal $g$ & 0.780 [0.655, 0.885] & 0.978 [0.926, 1.000] \\
Level $\log_{10}\Q$ at 15k & 0.788 [0.667, 0.889] & 0.974 [0.913, 1.000] \\
Post-learning rank drop & 0.792 [0.667, 0.900] & 0.935 [0.831, 1.000] \\
$\Q$-free set of eight statistics & 0.714 [0.579, 0.839] & 0.952 [0.866, 1.000] \\
Dormant fraction & 0.381 [0.250, 0.522] & 0.372 [0.186, 0.597] \\
\bottomrule
\end{tabular}
\end{table}

\paragraph{Split configurations.}\label{app:split} Walker2d and Ant at $r=4$ are the SAC configurations we ran to
80k in which the guard flags some seeds and leaves the others unflagged, so they are the configurations with seeds to
forecast. Across all SAC runs, the guard flags two Hopper runs, both at $r=16$, and no HalfCheetah or Humanoid run. At
30k, the phase map has two mixed cells: Ant at $r=8$ (3 of 6 seeds) and Hopper at $r=16$ (1 of 10;
Figure~\ref{fig:phase}a).

\subsection{Window sensitivity}\label{app:window} Table~\ref{tab:window}.
We set the window once, before the second Walker2d replicate, the Ant cohort and the live seeds existed.
For windows $(5\text{k},10\text{k}]$, $(5\text{k},20\text{k}]$ and $(10\text{k},20\text{k}]$ the Walker2d AUC ranges
from 0.754 to 0.811 and the Ant AUC from 0.909 to 1.000. The largest paired differences to $(5\text{k},15\text{k}]$,
$+0.031$ on Walker2d (DeLong 95\% CI $-0.026$ to $+0.089$) and $-0.069$ on Ant ($-0.161$ to $+0.022$), leave
room for differences near 0.1. With 2k logging the window $(5\text{k},10\text{k}]$ holds only 2 records for half the
Walker2d runs and for all Ant runs. Every DeLong interval for a difference to the paper's window includes zero, and the intervals are wide.

\begin{table}[h]\centering\small
\caption{LOSO AUC of $g$ for four windows, with 95\% bootstrap CIs, and the DeLong difference
to the paper's window $(5\text{k},15\text{k}]$ on the same runs with its 95\% CI.}
\label{tab:window}
\setlength{\tabcolsep}{3pt}
\resizebox{\linewidth}{!}{%
\begin{tabular}{@{}lcccc@{}}
\toprule
 & \multicolumn{2}{c}{Walker2d $r{=}4$ (64 runs, 31 flagged)} & \multicolumn{2}{c}{Ant $r{=}4$ (32 runs, 11 flagged)} \\
Window & AUC [95\% CI] & $\Delta$AUC [95\% CI] & AUC [95\% CI] & $\Delta$AUC [95\% CI] \\
\midrule
$(5\text{k},10\text{k}]$ & 0.754 [0.630, 0.861] & $-0.026$ [$-0.136$, $+0.083$] & 0.909 [0.775, 1.000] & $-0.069$ [$-0.161$, $+0.022$] \\
$(5\text{k},15\text{k}]$ & 0.780 [0.655, 0.885] & & 0.978 [0.926, 1.000] & \\
$(5\text{k},20\text{k}]$ & 0.811 [0.690, 0.905] & $+0.031$ [$-0.026$, $+0.089$] & 1.000 [1.000, 1.000] & $+0.022$ [$-0.016$, $+0.059$] \\
$(10\text{k},20\text{k}]$ & 0.762 [0.636, 0.870] & $-0.018$ [$-0.141$, $+0.106$] & 0.996 [0.974, 1.000] & $+0.017$ [$-0.022$, $+0.056$] \\
\bottomrule
\end{tabular}}
\end{table}

\subsection{Level versus slope (nested model)}\label{app:nested} Table~\ref{tab:nested}. The log level at 15k and
$g$ have Pearson correlation 0.988 on Walker2d and 0.998 on Ant. On Walker2d, the log level alone matches the slope,
and adding the slope to the level does not improve discrimination. A logistic model on
the \emph{raw} level scores lower (0.715) only because the raw level is heavy-tailed (median 1047, maximum
12,828); its model-free ranking AUC is 0.820, above the slope's 0.807 (difference $-0.013$, $p=0.55$). In
full-data likelihood-ratio tests the slope adds to the log level (LR 8.30, $p=0.004$), but with a
\emph{negative} coefficient given the level (standardized $-6.07$ against $+7.64$): at a fixed early level, a
faster slope lowers the fitted risk, so the pair encodes the early level; adding the log level to the slope gives
LR 12.37 ($p=4.4\times10^{-4}$). Given the level, the rate model of Section~\ref{sec:mechanism} predicts a positive
slope coefficient, since at a fixed early level a faster climb reaches the flag level sooner; the in-sample fit gives a
negative one ($p=0.004$), with the pair nearly collinear (Pearson 0.988). We therefore do not
claim that the rate rather than the level carries the forecast. On Ant both features are at ceiling and the test is not estimable (separation). The
DeLong differences in Table~\ref{tab:nested} compare nested models and are descriptive only.

\begin{table}[h]\centering\small
\caption{Walker2d $r{=}4$: LOSO AUC of the level, the slope and both (same folds).}
\label{tab:nested}
\begin{tabular}{@{}lcc@{}}
\toprule
Model & LOSO AUC & $\Delta$AUC vs.\ level only (DeLong 95\% CI, $p$) \\
\midrule
Slope $g$ & 0.780 & \\
Level $\log_{10}\Q$ at 15k & 0.788 & \\
Level + slope & 0.786 & $-0.002$ ($[-0.006, +0.002]$, 0.31) \\
Raw level $\Q$ at 15k & 0.715 & \\
Raw level + slope & 0.779 & $+0.065$ ($[+0.006, +0.123]$, 0.031) \\
\bottomrule
\end{tabular}
\end{table}

\subsection{Horizon, censoring and confirmation data}\label{app:horizon} Table~\ref{tab:horizon}. We fixed the
window on the first Walker2d replicate and collected replicate 2 and the live seeds afterwards. Harrell's concordance
index C \citep{harrell1982yield} is the fraction of comparable run pairs whose flag order the forecast ranks correctly:
a pair is concordant when the run flagged first has the larger early feature, with unflagged runs censored at 80k.
Harrell's C of $g$ stays close to its full-cohort value on replicate 2 and the live seeds
(0.746 against 0.783), and the AUC of $g$ itself lies between \iflnsection{}0.71\else{}0.72\fi{} and 0.88 in every Walker2d cohort of Table~\ref{tab:horizon}c. Labeling at $H=40$k, 50k, 60k, 70k and 80k gives Walker2d AUCs of
0.956, 0.912, 0.842, 0.784 and 0.780 (3 to 31 positives), while Ant stays at 0.97--1.00: on Walker2d, runs that
cross late start with early growth close to that of runs the guard never flags within 80k, so the AUC falls as the
horizon lengthens. Harrell's C scores the order of the flags instead and needs no horizon.
A truncated C that scores only the pairs whose earlier run flags by $\tau$ shows where the order is sharpest. On
Walker2d it is 0.962 (95\% CI 0.914--1.000) at $\tau=40$k, 0.933 at 50k, 0.857 at 60k and 0.803 at 70k, and at 80k
it equals the full C; on Ant it lies between 0.983 and 0.989 at every $\tau$. Ranked by $g$, the 3 Walker2d runs
flagged by 40k are 2nd, 4th and 6th of 64, and the 7 flagged by 50k are no lower than 16th; on Ant the 3 flagged by
40k are 1st, 2nd and 4th of 32, and the 7 flagged by 50k no lower than 8th. Among flagged runs,
faster early growth goes with an earlier flag: Spearman $\rho=-0.64$ ($p=1.1\times10^{-4}$) on Walker2d and $-0.96$
($p=1.9\times10^{-6}$) on Ant (Figure~\ref{fig:forecast}b).

\begin{table}[h]\centering\footnotesize
\caption{(a) LOSO AUC of $g$ when the label is ``flagged by $H$''. (b) Harrell's C (95\% bootstrap CI) against
the flag step. (c) Walker2d by cohort: AUC of $g$ itself (95\% bootstrap CI) and LOSO AUC; the live row
applies a model fit on all 64 Walker2d seeds.}
\label{tab:horizon}
\setlength{\tabcolsep}{4pt}
\begin{tabular}{@{}lccccc@{}}
\toprule
(a) $H$ & 40k & 50k & 60k & 70k & 80k \\
\midrule
Walker2d $r{=}4$: positives, AUC & 3, 0.956 & 7, 0.912 & 16, 0.842 & 22, 0.784 & 31, 0.780 \\
Ant $r{=}4$: positives, AUC & 3, 0.966 & 7, 0.983 & 8, 1.000 & 11, 0.978 & 11, 0.978 \\
\bottomrule
\end{tabular}

\vspace{4pt}
\begin{tabular}{@{}lcc@{}}
\toprule
(b) Cohort & C of $g$ & C of level \\
\midrule
Walker2d $r{=}4$ (64) & 0.783 [0.690, 0.864] & 0.802 [0.711, 0.876] \\
Ant $r{=}4$ (32) & 0.983 [0.958, 1.000] & 0.979 [0.953, 1.000] \\
Walker2d replicate 2 and live seeds (60) & 0.746 [0.651, 0.829] & 0.758 [0.662, 0.841] \\
\bottomrule
\end{tabular}

\vspace{4pt}
\begin{tabular}{@{}lcc@{}}
\toprule
(c) Walker2d cohort & AUC of $g$ & LOSO AUC \\
\midrule
Replicate 1 (window fixed here) & 0.875 [0.722, 0.992] & 0.827 \\
Replicate 2 & 0.723 [0.531, 0.891] & 0.629 \\
Live seeds & 0.749 [0.545, 0.920] & 0.749 (fit on 64) \\
Replicate 2 and live seeds & 0.742 [0.607, 0.857] & 0.691 \\
\iflnsection
LayerNorm study, plain arm (Sec.~\ref{sec:ln}) & 0.714 [0.518, 0.875] & 0.651 \\
\fi
\bottomrule
\end{tabular}
\end{table}

\paragraph{A return-based endpoint.} Calling a run failed when its mean return over its last 10k logged steps
is below half of its best 10k-window mean labels 1 of 64 Walker2d runs; with 4k windows it labels 8 (7 of them
unflagged), and $g$ does not predict them (LOSO AUC 0.426). Within 80k, return rarely collapses, so a return
label cannot replace the flag here.

\subsection{Collapse features within the split configurations}\label{app:rankfeat} Read at the 15k decision step,
the post-learning rank drop ($1-$ rank at the last record up to 15k over the maximum rank after the first gradient
step) has LOSO AUC 0.792 (95\% CI 0.667--0.900) on Walker2d and 0.935 (0.831--1.000) on Ant, and Spearman correlation
$+0.59$ and $+0.95$ with $g$. The rank drop from the all-run peak reaches 0.671 and 0.957, and the dormant
fraction 0.381 and 0.372. The dormant fraction's LOSO AUC falls below 0.5 because leave-one-out fitting pushes an
uninformative statistic below chance; its AUC itself, with no fit, is 0.530 on Walker2d and 0.431 on Ant. Our
analysis script \texttt{r15\_rank\_measures.py} (available from the corresponding author on request) computes two related rank drops: it reads the median of the last three
records up to 15k and takes the post-learning peak over the records after step 5k.

\subsection{Timing across UTD ratios}\label{app:whenr} Table~\ref{tab:whenr}. Flag steps per cell (SAC,
no clipping, each run counted once; the event-timing runs replace the originals they replay;
at $r=16$ the cells also hold 4 Walker2d and 3 Ant runs of a multi-task cohort run to 60k, which
Tables~\ref{tab:collapse} and~\ref{tab:rate} do not include),
and the ``when'' forecast with the window scaled to 40k gradient updates, $(5\text{k}, 5\text{k}+40\text{k}/r]$,
on the flagged runs whose flag comes after the window. With the scaled window the timing forecast also works where
the guard flags nearly every seed: 1.65k against 2.92k on Walker2d at $r=8$ ($p=0.039$), 1.66k against 3.38k
on Ant at $r=16$ ($p=0.019$), 1.01k against 1.48k on Walker2d at $r=16$ ($p=0.76$). The 15k decision step gives no
lead time there: 12 of 14 Walker2d and 8 of 11 Ant flags at $r=16$ come at or before 15k, and Walker2d flags at
$r=8$ come from 17.4k on (median 22.1k). The decision step must follow the UTD ratio; at $r=4$ the Walker2d flag
is still 18.1k to 65.0k steps away at 15k.

\begin{table}[h]\centering\footnotesize
\caption{Flag timing and the LOSO forecast of the flag step by UTD ratio (mean absolute error, MAE, of a
linear regression on $g$ against a constant LOSO-median forecast; paired Wilcoxon $p$). At $r{=}16$ the cells
include a multi-task cohort (4 Walker2d and 3 Ant runs, $H=60$k) besides the phase-map runs.}
\label{tab:whenr}
\setlength{\tabcolsep}{4pt}
\begin{tabular}{@{}lcccccc@{}}
\toprule
Cell & Flagged & Earliest, median flag & Flags $\le$15k & Window & MAE $g$ / constant & $p$ \\
\midrule
Walker2d $r{=}4$ (80k) & 31/64 & 33.1k, 59.8k & 0 & $(5\text{k},15\text{k}]$ & 7.85k / 11.31k & 0.048 \\
Ant $r{=}4$ (80k) & 11/32 & 35.7k, 47.3k & 0 & $(5\text{k},15\text{k}]$ & 3.13k / 9.43k & 0.005 \\
Walker2d $r{=}8$ & 8/8 & 17.4k, 22.1k & 0 & $(5\text{k},10\text{k}]$ & 1.65k / 2.92k & 0.039 \\
Ant $r{=}8$ & 3/6 & 17.8k, 18.4k & 0 & & too few & \\
Walker2d $r{=}16$ & 14/14 & 9.1k, 10.6k & 12 & $(5\text{k},7.5\text{k}]$ & 1.01k / 1.48k & 0.76 \\
Ant $r{=}16$ & 11/11 & 9.0k, 12.8k & 8 & $(5\text{k},7.5\text{k}]$ & 1.66k / 3.38k & 0.019 \\
\bottomrule
\end{tabular}
\end{table}

\subsection{Calibration}\label{app:ece} Table~\ref{tab:ece}. The slope-only model is the one the controller uses.
A 14-feature model, the slopes and last values of seven channels (effective rank, dormant fraction, return,
weight norm, raw $\Q$, TD loss and gradient norm), which include the raw $\Q$ slope and level but not the
$\log_{10}\Q$ slope, has a similar calibration error and a lower LOSO AUC (0.736 on Walker2d, 0.952 on Ant). The
percentile CIs of the ECE lie mostly above the point estimate, as usual for a statistic bounded below by zero.

\begin{table}[h]\centering\small
\caption{Expected calibration error of LOSO probabilities (5 equal-width bins; 95\% bootstrap CI for the
slope-only model; equal-mass bins in the last row).}
\label{tab:ece}
\resizebox{\linewidth}{!}{%
\begin{tabular}{@{}lccc@{}}
\toprule
Model & Walker2d & Ant & Pooled ($n{=}96$) \\
\midrule
Slope only (forecast and controller) & 0.094 [0.049, 0.205] & 0.078 [0.058, 0.174] & 0.065 [0.044, 0.153] \\
14 features incl.\ raw $\Q$ (seven channels) & 0.087 & 0.115 & 0.072 \\
Slope only, equal-mass bins & 0.135 & 0.100 & 0.087 \\
\bottomrule
\end{tabular}}
\end{table}

\section{A Rate Model of Value Growth}\label{app:rate}
\paragraph{Model.} For run $i$ of one configuration and steps $t$ after the warm-up $t_0=5$k,
\begin{equation*}
\log_{10}\Q_i(t)=a_i+g_i\,h(t),
\end{equation*}
with a run-specific start $a_i$ and rate $g_i>0$, and a growth profile $h$ shared by the runs of the configuration:
continuous and strictly increasing, with $h(t_0)=0$, and scaled so that its least-squares slope over the early window
$(t_0,t_d]$, $t_d=15$k, is one. The early-growth signal of run $i$ is then $g_i$. For a flagged run, the guard fires at the first $t$ with
$\log_{10}\Q_i(t)\ge c=\log_{10}\tau$, and $a_i<c$ for every run. This rule idealizes the flagged runs, which flag in a
narrow band of $\Q$ (below); reaching the band does not by itself flag a run, so the proposition orders the flagged
runs, which is what the Spearman correlations of Table~\ref{tab:rate} measure. A bootstrapped update that multiplies the values
by a constant ratio per step gives a linear $h$; linearized analyses of temporal-difference learning tie divergence to
such updates \citep{achiam2019characterizing}. For multilayer critics trained offline with Adam, SEEM derives growth
polynomial in the number of updates, of degree equal to the network's depth, so that $\log\Q$ grows with the logarithm
of the update count \citep{yue2023seem}; a profile of that kind is concave, the case of part (c) below.

\paragraph{Proposition (order and early extrapolation).} Under the model:
(a) $T_{{\rm flag},i}=h^{-1}\big((c-a_i)/g_i\big)$.
(b) For every $h$, $T_{{\rm flag},i}<T_{{\rm flag},j}$ exactly when $(c-a_i)/g_i<(c-a_j)/g_j$; in particular, two runs
with the same start flag in the order of their rates.
(c) If $h$ is concave, the straight line through run $i$'s value at $t_d$ with slope $g_i$ reaches $c$ at
$\hat T_i=t_d+\big(c-\log_{10}\Q_i(t_d)\big)/g_i\le T_{{\rm flag},i}$: extending the early line lands early.

\paragraph{Proof.} (a) $\log_{10}\Q_i(t)\ge c$ exactly when $h(t)\ge(c-a_i)/g_i$, and $h$ is continuous and strictly
increasing, so the first such $t$ is $h^{-1}\big((c-a_i)/g_i\big)$. (b) $h^{-1}$ is strictly increasing. With $a_i=a_j<c$,
$(c-a)/g$ falls as $g$ rises, so the faster run flags first. (c) The least-squares slope of $h$ over the window's
records is a weighted average, with nonnegative weights, of the slopes of $h$ between consecutive records. For concave
$h$ these slopes do not increase, so the slope of any chord of $h$ that starts at $t_d$ is at most the window's slope,
which is one: $h(t)-h(t_d)\le t-t_d$ for $t>t_d$. Hence $\log_{10}\Q_i(t)\le\log_{10}\Q_i(t_d)+g_i(t-t_d)$, the line lies
on or above the curve after $t_d$, and it reaches $c$ no later than the curve does. $\square$

The starts are close. Take run $i$'s
least-squares line of $\log_{10}\Q$ over the early window: its value at 5k has standard deviation 0.110 across the 64
Walker2d runs and 0.033 across the 32 Ant runs, against 0.399 and 0.256 for its rise over the window, $10g_i$. The
level at 15k then ranks the runs nearly as $g$ does (Spearman correlation 0.965 and 0.995).
Spread in the starts and in the flag level is too small to account for the flagged runs that flag out of rate order
on Walker2d. With each flagged run's measured start $a_i$ and the common level $c$, the order $(c-a_i)/g_i$ that part
(b) implies is nearly the order of $g$ (Spearman correlation $-0.987$), and it matches the flag order no better than
$g$ does: its Spearman correlation with the flag step is $+0.644$, and $+0.643$ with each run's own flag level,
against $-0.639$ for $g$. The model thus gets the direction of the flag order right on Walker2d at $r=4$, and the
order observed there is looser than the one it implies. We did not test what produces the remaining disorder;
profiles that differ between runs and error in the early slope are two untested possibilities. On Ant the model's order matches the data ($+0.945$, against $-0.964$ for $g$).

\paragraph{Tests.} Table~\ref{tab:rate} lists the tests, on the split configurations (80k) and the phase-map cells
with at least five flagged runs (30k). P1--P6 test the linear $h$ of constant-ratio growth and related hypotheses
against the criteria in the caption; part (c) of the proposition covers a concave $h$, the bend that P1 reports. P2's correlations on the split configurations also appear in Appendix~\ref{app:horizon}; its new test is
the phase-map cells.
The guard fires near a common level: the batch-mean $\Q$ of the flag record has median $2.01\times10^5$ (range
$1.01\times10^5$--$2.73\times10^5$) over the 31 flagged Walker2d runs at $r=4$ and $3.24\times10^5$
($2.27\times10^5$--$4.23\times10^5$) over the 11 flagged Ant runs. Reaching this band is not sufficient for a flag:
18 of the 33 unflagged Walker2d runs end the budget at or above the lowest flag level, and 3 of them above the median
flag level, while no unflagged Ant run reaches the lowest Ant flag level. The curve bends downward: the quadratic term
of a fit to $\log_{10}\Q$ from 5k to the flag is negative in all 72 flagged runs of the split configurations and the
30k phase map, and P1 falls short of its $R^2$ criterion on Walker2d at $r=4$, the cohort with the longest climb to the flag
(median 38 records).
As part (c) predicts, the fit-free extrapolation $\hat T$ lands early, by a median 31.95k steps on Walker2d and 16.15k
on Ant, and so loses to the constant median forecast (P3); the order it rests on holds in all three tested cohorts
whose flags all come after the early window (P2).

\begin{table}[h]\centering\footnotesize
\caption{Tests of the rate model and related hypotheses. (i) Consequences of the model. One further consequence is
not borne out: at a fixed early level the model implies a positive coefficient on $g$, and the fit on the 80k label
gives a negative one (Appendix~\ref{app:nested}). P2, part (b): Spearman
correlation of $g$ with the flag step among flagged runs, 95\% bootstrap CI below zero, in the three tested cohorts
whose flags all come after the early window. At $r=16$, $g$ of a run flagged before 15k is read over $(5\text{k},\text{flag}]$, which
builds part of the correlation in; 5 of the 8 flagged Ant runs and 8 of the 10 flagged Walker2d runs flag at or
before 15k. There the correlation is $-0.976$ (Ant) and $-0.939$ (Walker2d), and over the fixed window
$(5\text{k},7.5\text{k}]$, which ends before every flag, $-0.929$ (2 records per run) and $-0.794$, both CIs below zero;
we report these two cells as descriptive. (c), part (c): median signed error of the fit-free $\hat T$. (ii) Related
hypotheses. P3: $\hat T$ beats the constant median forecast (CI of the difference in mean absolute error below zero),
which part (c) does not predict. P1, a linear $h$: median $R^2$ of a
straight line in $\log_{10}\Q$ from 5k to the flag $\ge0.90$, better than a line in $\Q$ in a majority of runs,
Wilcoxon $p<0.05$. P4: the cell median of $g$ rises with $r$ in every task (30k phase map).\iflnsection{} P5: LayerNorm
lowers $g$ (paired).\fi{} P6: on Walker2d at $r=4$, $g$ is unrelated to the collapse measures at 15k (both CIs include
zero).}
\label{tab:rate}
\setlength{\tabcolsep}{3pt}
\begin{tabular}{@{}lp{8.9cm}l@{}}
\toprule
 & Result & Verdict \\
\midrule
\multicolumn{3}{@{}l}{\emph{(i) Consequences of the model}} \\
P2 & $-0.639$ [$-0.820$, $-0.340$] (Walker2d $r{=}4$), $-0.964$ [$-1.000$, $-0.822$] (Ant $r{=}4$), $-0.905$
[$-1.000$, $-0.595$] (Walker2d $r{=}8$) & holds in all three \\
(c) & $\hat T$ minus the flag step, median: $-31.95$k (Walker2d $r{=}4$), $-16.15$k (Ant $r{=}4$) & lands early \\
\multicolumn{3}{@{}l}{\emph{(ii) Related hypotheses}} \\
P3 & MAE of $\hat T$ minus MAE of the constant: $+22.11$k [$+16.97$, $+26.76$] on Walker2d, $+7.67$k [$+3.11$, $+11.64$]
on Ant (MAE of $\hat T$ 33.42k and 17.10k) & fails \\
P1 & Walker2d $r{=}4$: $R^2$ 0.891 [0.883, 0.907], log better in 22/31, $p=0.041$. Ant $r{=}4$: 0.975, 11/11.
Ant $r{=}16$: 0.987, 8/8. Walker2d $r{=}8$: 0.948, 8/8. Walker2d $r{=}16$: 0.968, 10/10 & falls just short on Walker2d $r{=}4$ \\
P4 & Median $g$ at $r=2$ and $r=16$: Walker2d 0.116 and 0.631, Ant 0.080 and 0.463 (run-level Spearman with $r$ $+0.875$,
$+0.826$); Humanoid 0.132 and 0.017 ($-0.968$); HalfCheetah $-0.194$ and Hopper $+0.344$, both CIs including zero & fails \\
\iflnsection
P5 & $g$ lower with LayerNorm in 31/32 pairs, median difference $-0.0419$ [$-0.0553$, $-0.0310$] & holds \\
\fi
P6 & Spearman of $g$ with the post-learning rank drop $+0.591$ [$+0.392$, $+0.748$], with the dormant fraction $+0.146$
[$-0.112$, $+0.386$]; Ant: $+0.948$ and $-0.142$ & fails \\
\bottomrule
\end{tabular}
\end{table}

\paragraph{What the failures show.} P3 fails in the direction part (c) predicts, so the model keeps the order and gives
up the timing: turning $g$ into a flag step takes a fitted map, which the LOSO regression on $g$ supplies (7.85k and
3.13k against 11.31k and 9.43k for the constant). P4 fails because the rate rises with $r$ in Walker2d and Ant but falls in
Humanoid and stays flat in HalfCheetah, so we read the rate within one configuration only. P6 fails on the rank drop, which rises with $g$, while the dormant fraction stays unrelated to it
(Section~\ref{sec:mechanism}).

\section{Growth-Abort Controller}\label{app:ctrl}
\paragraph{Slope used by the controller.} Algorithm~\ref{alg:ctrl} computes the slope of $\ln\Q$ (not
$\log_{10}\Q$) over the records with at least one gradient step and step $\le t_d$, the quantity the live
trainer computes. On every run these are the records of $g$'s window, so the controller slope is exactly
$\ln 10$ times $g$, and $\theta=3.7681\times10^{-4}$ per step equals $g=0.164$ per 1k steps. The controller
aborts a run when its slope is strictly greater than $\theta$; the in-sample optimum sits on one calibration seed's slope.

\begin{figure}[h]\refstepcounter{algo}\label{alg:ctrl}
\noindent\fbox{\parbox{0.975\linewidth}{\small
\textbf{Algorithm~\thealgo: the growth-abort controller.} Input: training seeds with logged $\Q$ and flag
steps; decision step $t_d=15$k; horizon $H$.
\begin{enumerate}\setlength{\itemsep}{0pt}\setlength{\parskip}{0pt}
\item On the training seeds, compute each run's slope of $\ln\Q$ over the logged records up to $t_d$ that
follow the first gradient step, and choose the $\theta$ that maximizes net steps saved (a true abort saves
$T_{\rm flag}-t_d$; a false abort costs $H$).
\item Run each test seed to $t_d$, compute its slope from its own log, and abort it if the slope is strictly
greater than $\theta$; otherwise continue to $H$.
\item Out of sample: apply steps 1--2 per fold of an 8-fold split over seeds. Live: fix $\theta$ in
advance and run step 2 inside the trainer.
\end{enumerate}}}
\end{figure}

\begin{figure}[h]\centering
\begin{minipage}[c]{0.58\linewidth}\centering
\includegraphics[width=\linewidth]{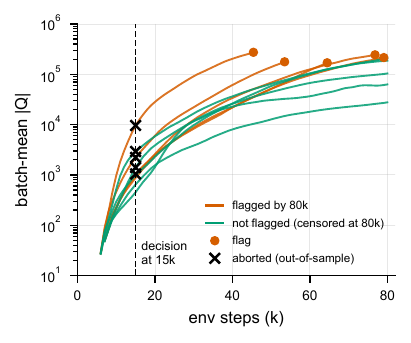}
\end{minipage}\hfill
\begin{minipage}[c]{0.39\linewidth}\raggedright
\caption{\textbf{The controller decides at 15k, long before any flag.} Ten Walker2d $r{=}4$ runs, drawn at random
within each outcome type: 3 true aborts, 2 missed flags, 2 false aborts and 3 correct continues. A cross marks an
out-of-sample abort at 15k; a dot marks the flag. Table~\ref{tab:ledger} gives the savings of every
cohort.}\label{fig:decide}
\end{minipage}
\end{figure}

\paragraph{Accounting.} Table~\ref{tab:ledger} extends Table~\ref{tab:ctrl} with every row of the per-seed
ledger, including the charge of only the lost remainder $H-t_d$ per false abort and the TD3 cohort. Runs flagged at or before $t_d$ leave the decision set
(one TD3 run, flagged at 12,086). The retrospective row sweeps a threshold on the LOSO logistic risk; its best
point, at the 55th percentile, saves 364,109 steps under the full-horizon charge (469,109 under the remainder
charge). Over 200 random 8-fold splits, the Walker2d net saving has median 444,109, which is also its maximum
(5th percentile 204,109, minimum 72,581). The best threshold in hindsight saves 524,109 steps on Walker2d; it sits
exactly on one seed's slope ($\theta=3.7681\times10^{-4}$ in $\ln\Q$ per step, 0.164 in $\log_{10}\Q$ per 1k steps
as in Figure~\ref{fig:forecast}b), and $\theta=3.768\times10^{-4}$ would abort that seed too (444,109).

\begin{table}[h]\centering\footnotesize
\caption{The full per-seed ledger (decision at 15k, $H=80$k). ``Net'' charges a false abort $H$,
``alt.'' charges $H-t_d$. For the live cohorts and TD3 we report the net saving and the oracle.}
\label{tab:ledger}
\setlength{\tabcolsep}{2.5pt}
\begin{tabular}{@{}lrcccrrrr@{}}
\toprule
Cohort & $n$ & TP/FP/FN & Prec. & Rec. & Gross & Net & Net, alt. & Oracle \\
\midrule
\multicolumn{9}{@{}l}{\emph{SAC Walker2d $r{=}4$}} \\
\quad oracle & 64 & 31/0/0 & 1.00 & 1.00 & 1{,}391{,}847 & 1{,}391{,}847 & 1{,}391{,}847 & 1{,}391{,}847 \\
\quad best threshold in hindsight & 64 & 22/5/9 & 0.81 & 0.71 & 924{,}109 & 524{,}109 & 599{,}109 & 1{,}391{,}847 \\
\quad retrospective sweep, best point & 64 & 22/7/9 & 0.76 & 0.71 & 924{,}109 & 364{,}109 & 469{,}109 & 1{,}391{,}847 \\
\quad K-fold ($K{=}8$) & 64 & 22/6/9 & 0.79 & 0.71 & 924{,}109 & 444{,}109 & 534{,}109 & 1{,}391{,}847 \\
\quad leave-one-seed-out & 64 & 22/6/9 & 0.79 & 0.71 & 924{,}109 & 444{,}109 & 534{,}109 & 1{,}391{,}847 \\
\multicolumn{9}{@{}l}{\emph{SAC Walker2d $r{=}4$, live, threshold fixed in advance}} \\
\quad 21 paired fresh seeds & 21 & -- & -- & -- & -- & 130{,}085 & -- & 512{,}263 \\
\quad all 28 fresh seeds & 28 & -- & -- & -- & -- & 252{,}389 & -- & 803{,}515 \\
\multicolumn{9}{@{}l}{\emph{SAC Ant $r{=}4$}} \\
\quad best threshold in hindsight & 32 & 10/0/1 & 1.00 & 0.91 & 327{,}386 & 327{,}386 & 327{,}386 & 372{,}819 \\
\quad K-fold ($K{=}8$) and LOSO & 32 & 10/1/1 & 0.91 & 0.91 & 327{,}386 & 247{,}386 & 262{,}386 & 372{,}819 \\
\multicolumn{9}{@{}l}{\emph{TD3 Walker2d $r{=}16$}} \\
\quad K-fold ($K{=}8$) & 30 & -- & -- & -- & -- & $-$81{,}365 & -- & 232{,}660 \\
\bottomrule
\end{tabular}
\end{table}

\subsection{Threshold sensitivity, calibration cost and other stopping rules}\label{app:alt}
Table~\ref{tab:alt}. All rows use the ledger above and, for the out-of-sample rows, the same 8-fold split.
In (a), $\theta^*$ is the in-sample best threshold of each cohort ($3.7680916\times10^{-4}$ on Walker2d,
$3.0398774\times10^{-4}$ on Ant). In (b), a threshold is fit on $n$ calibration seeds drawn at random from the
64 Walker2d seeds and applied to the other $64-n$ (500 draws). The profitable band of $\theta$ is narrow: on
Walker2d $0.95\theta^*$ loses 152,449 steps and $1.2\theta^*$ loses 4,307, and Ant's $\theta^*$ is 0.807 times
Walker2d's. The 64 labeled calibration runs cost 4,496,847 steps, while a threshold fit on 56 of them saves a
median 7,702 steps per held-out run, so calibration pays for itself only after 584 further runs of the configuration.

\paragraph{Other stopping rules (Table~\ref{tab:ctrl}b).} Each rule aborts when its statistic at 15k is above a
threshold fit on the training folds (below, for return). The two label-free rules need no threshold fitting: the
median stopping rule aborts a run whose return at 15k is below the median of the other runs
\citep{golovin2017vizier}, and the cohort rule aborts a run whose slope is above the other runs' 75th percentile.
We chose that percentile without separate data, so Table~\ref{tab:alt}c sweeps it: from the 60th to the 85th
percentile the cohort rule saves compute on both split configurations, and it loses at the 50th on Ant and at the
90th on Walker2d. The rule assumes that the guard will flag a quarter or more of the cohort. On the $r=4$ controls, which
the guard never flags, it aborts 4 of the 12, 16 and 12 runs of each task, and its 12 false aborts lose 960,000
steps, 24,000 per run. Table~\ref{tab:rules} gives the net saving of every rule of Table~\ref{tab:ctrl}b with
its TP/FP/FN counts.

\begin{table}[h]\centering\footnotesize
\caption{Table~\ref{tab:ctrl}b in full: net environment steps saved by each abort rule read at 15k on the same
8-fold split, with TP/FP/FN (true aborts, false aborts, missed flags).}
\label{tab:rules}
\begin{tabular}{@{}lcc@{}}
\toprule
Abort rule, read at 15k & Walker2d $r{=}4$ & Ant $r{=}4$ \\
\midrule
Early-growth signal $g$ (the controller) & 444{,}109 (22/6/9) & 247{,}386 (10/1/1) \\
Level $\log_{10}\Q$ & 332{,}891 (20/6/11) & 247{,}386 (10/1/1) \\
Post-learning rank drop & 438{,}389 (20/5/11) & 150{,}515 (8/1/3) \\
Dormant fraction & $-$261{,}089 (1/4/30) & $-$80{,}000 (0/1/11) \\
Low return & 103{,}027 (6/2/25) & $-$80{,}000 (0/1/11) \\
Median stopping rule on return, label-free & $-$588{,}012 (16/16/15) & $-$705{,}514 (5/11/6) \\
$g$ above the cohort's 75th percentile, label-free & 150{,}898 (13/4/18) & 230{,}515 (8/0/3) \\
\bottomrule
\end{tabular}
\end{table}

\paragraph{What an abort gives up.} The ledger counts a flagged run as lost, because the guard stops it before its
budget ends, so the ledger values a run only if the run completes its budget. With the in-sample threshold, the 27
aborted Walker2d runs have median best logged return 453.6, reached after 15k in all 27, against 444.1 for the 37
continued runs (Mann--Whitney $p=0.97$); under the K-fold decisions, 466.8 against 440.6 ($p=0.93$). A practitioner
who keeps each run's best checkpoint therefore gives up, with each abort, a policy comparable to those of the runs
the controller continues. Across the 64 runs, $g$ and the best return are nearly uncorrelated (Spearman $+0.14$,
$p=0.25$), so early $\Q$ growth carries almost no information about how fast a run learns.

\begin{table}[h]\centering\footnotesize
\caption{(a) Net saving against $\theta/\theta^*$ (in sample). (b) Net saving per held-out run when $\theta$ is
fit on $n$ calibration seeds (median and 5--95\% range over draws; share of draws that lose). (c) Net saving of the
label-free cohort rule against its percentile (the rule aborts a run whose slope is above that percentile of the
other runs' slopes). The other abort rules are in Table~\ref{tab:ctrl}b.}
\label{tab:alt}
\setlength{\tabcolsep}{3pt}
\begin{tabular}{@{}lccccccccc@{}}
\toprule
(a) $\theta/\theta^*$ & 0.80 & 0.85 & 0.90 & 0.95 & 1.00 & 1.05 & 1.10 & 1.15 & 1.20 \\
\midrule
Walker2d (k) & $-688$ & $-448$ & $-368$ & $-152$ & 524 & 431 & 151 & 140 & $-4$ \\
Ant (k) & $-667$ & $-347$ & $-187$ & $-27$ & 327 & 231 & 231 & 156 & 94 \\
\bottomrule
\end{tabular}

\vspace{4pt}
\resizebox{\linewidth}{!}{%
\begin{tabular}{@{}lccccc@{}}
\toprule
(b) Calibration seeds $n$ & 8 & 16 & 32 & 48 & 56 \\
\midrule
Median per held-out run & 2{,}531 & 3{,}116 & 4{,}901 & 5{,}982 & 7{,}702 \\
5--95\% & [$-13{,}329$, 7{,}518] & [$-7{,}770$, 8{,}504] & [$-4{,}971$, 11{,}204] & [$-8{,}620$, 17{,}253] & [$-14{,}443$, 24{,}523] \\
Share of draws with net $\le0$ & 0.34 & 0.23 & 0.19 & 0.24 & 0.29 \\
\bottomrule
\end{tabular}}

\vspace{4pt}
\begin{tabular}{@{}lccccccc@{}}
\toprule
(c) Cohort percentile & 50 & 60 & 70 & 75 & 80 & 85 & 90 \\
\midrule
Walker2d (k) & 243 & 398 & 282 & 151 & 233 & 140 & $-4$ \\
Ant (k) & $-27$ & 213 & 327 & 231 & 196 & 156 & 94 \\
\bottomrule
\end{tabular}
\end{table}

\subsection{Live deployment}\label{app:live} We analyze 21 fresh Walker2d seeds, none of them used to set $\theta$,
each run twice with identical settings: once without control to 80k, and once with the controller active in the
trainer ($\theta=3.7681\times10^{-4}$ in $\ln\Q$ per step). The uncontrolled cohort has 28 fresh runs, including
these 21. On all 21 pairs the controlled run's slope at 15k equals its twin's, and the continued pairs agree at every
logged record (the guard flags seed 103 at 77,178 in both arms), so on one machine runs with identical settings are
deterministic given the seed. Table~\ref{tab:live} lists the live errors on the 21 pairs. The true aborts have slopes 1.8\% to 55.8\%
above $\theta$, and the guard flagged them at 43,543 to 72,196 steps; the true negatives sit 27.1\% to 2.3\% below
$\theta$ and end at a median $\Q$ of $1.08\times10^5$.

\begin{table}[h]\centering\footnotesize
\caption{Live errors on the 21 paired seeds. Return: mean of the last 10 logged returns of the
uncontrolled twin. Late growth: slope of $\log_{10}\Q$ per 10k steps over the last 20k of an unflagged
twin.}
\label{tab:live}
\setlength{\tabcolsep}{4pt}
\begin{tabular}{@{}rlrrcl@{}}
\toprule
Seed & Error & Slope vs.\ $\theta$ & Return & $\Q$ at 15k / 40k / end & Twin \\
\midrule
115 & false abort & $+2.4\%$ & 195 & $1.33{\times}10^3$ / $7.77{\times}10^4$ / $1.31{\times}10^5$ & late growth 0.042 \\
126 & false abort & $+40.3\%$ & 208 & $3.54{\times}10^3$ / $1.57{\times}10^4$ / $4.67{\times}10^4$ & late growth 0.017 \\
127 & false abort & $+1.8\%$ & 365 & $1.21{\times}10^3$ / $3.07{\times}10^4$ / $1.46{\times}10^5$ & late growth 0.123 \\
131 & false abort & $+7.4\%$ & 204 & $1.44{\times}10^3$ / $3.54{\times}10^4$ / $1.63{\times}10^5$ & late growth 0.097 \\
103 & missed flag & $-5.5\%$ & 374 & $1.08{\times}10^3$ / $2.82{\times}10^4$ / $2.13{\times}10^5$ & flagged at 77,178 \\
\bottomrule
\end{tabular}
\end{table}

\subsection{Transfer across configurations}\label{app:transfer} Table~\ref{tab:transfer}: a threshold
calibrated on one cohort (the rule above, in sample) and applied unchanged to others. The threshold transfers
poorly across tasks. The Walker2d threshold applied to Ant aborts only 4 runs (precision 1.00, recall 0.36) and
saves 93,990 steps, against 247,386 with Ant's own; the Ant threshold applied to Walker2d aborts 57 of 64 runs and
loses 688,153. We did not test transfer across UTD ratios within a task.

\begin{table}[h]\centering\small
\caption{Abort thresholds applied across configurations (decision at 15k; net charges a false abort the
full horizon). For the 28 fresh seeds and for TD3 we report the saving.}
\label{tab:transfer}
\begin{tabular}{@{}lrrccr@{}}
\toprule
Target cohort & $n$ & Aborts & Prec. & Rec. & Net \\
\midrule
\multicolumn{6}{@{}l}{\emph{Threshold from SAC Walker2d $r{=}4$ ($3.7681\times10^{-4}$)}} \\
SAC Ant $r{=}4$ (80k) & 32 & 4 & 1.00 & 0.36 & 93{,}990 \\
SAC Walker2d $r{=}4$, 28 fresh seeds & 28 & -- & -- & -- & 252{,}389 \\
SAC Ant $r{=}8$ (30k) & 6 & 4 & 0.75 & 1.00 & $-$12{,}674 \\
SAC Hopper $r{=}16$ (30k) & 10 & 5 & 0.20 & 1.00 & $-$116{,}543 \\
TD3 Walker2d $r{=}16$ & 30 & -- & -- & -- & $-$324{,}300 \\
\multicolumn{6}{@{}l}{\emph{Threshold from SAC Ant $r{=}4$ ($3.040\times10^{-4}$)}} \\
SAC Walker2d $r{=}4$ & 64 & 57 & 0.54 & 1.00 & $-$688{,}153 \\
SAC Walker2d $r{=}4$, 28 fresh seeds & 28 & -- & -- & -- & 3{,}515 \\
TD3 Walker2d $r{=}16$ & 30 & -- & -- & -- & $-$948{,}443 \\
\bottomrule
\end{tabular}
\end{table}

\section{Collapse and Resets}\label{app:collapse}
\begin{figure}[t]\centering\includegraphics[width=\linewidth]{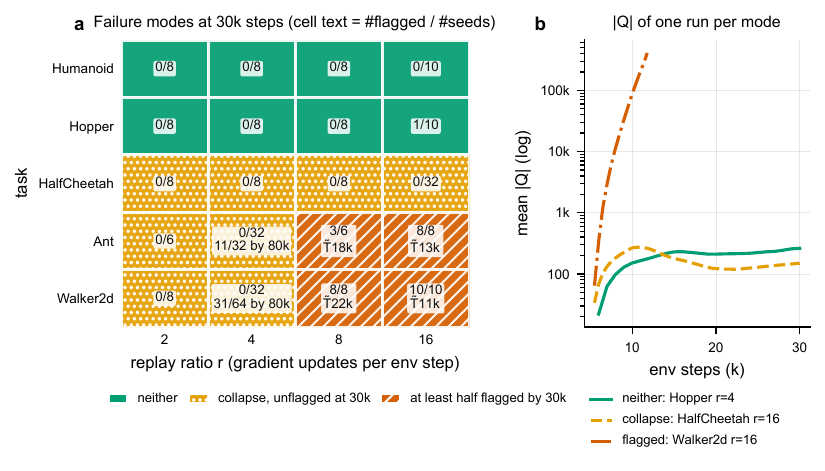}
\caption{\textbf{Collapse occurs without divergence.} (a) Phase map at 30k steps: we mark a cell \emph{mostly
flagged} when the guard flags at least half its seeds, else collapsed (Section~\ref{sec:setup}) or neither; all four
mostly flagged cells also meet the collapse rule (Appendix~\ref{app:collapse}). Cells print flagged/seeds and the
median flag step $\tilde T$; the Walker2d and Ant cells at $r{=}4$ also print the flag counts of the 80k forecast
cohorts (Table~\ref{tab:design}). (b) $\Q$ of one run per mode.}\label{fig:phase}
\end{figure}

\paragraph{What the collapse label measures.}
Table~\ref{tab:collapse} reprints, per phase-map cell, the two quantities the label uses and the drop
measured from the post-learning peak. In 204 of 234 runs the peak effective rank is the untrained
critic's. In HalfCheetah the post-learning drop is small (0.05--0.10) while about three quarters of the units
are dormant; in Walker2d and Ant the rank keeps falling after learning starts, and more so at higher $r$. Across
the 20 cells, the post-learning drop has Spearman correlation 0.83 with an indicator of Walker2d and Ant, the two
tasks in which the guard flags runs at $r\le8$ ($p=5.4\times10^{-6}$), and 0.42 with the 30k flag rate ($p=0.064$);
the final dormant fraction has 0.42 ($p=0.062$) and 0.20 ($p=0.39$). The indicator leaves out Hopper, whose flagged
runs all come at $r=16$ (Appendix~\ref{app:split}) and whose post-learning drop is 0.01 in every cell. Every cell
that Figure~\ref{fig:phase} draws as mostly flagged (Walker2d and Ant at $r\in\{8,16\}$) also meets the collapse rule;
the figure draws mostly flagged first. Collapse therefore accompanies divergence in the mostly flagged cells, but
the two do not always go together: HalfCheetah collapses without divergence, and the Hopper cell at $r=16$, in
which the guard flags a run, does not meet the collapse rule.

\paragraph{How fragile the label is.} Ant's median final dormant fraction, 0.507 to 0.517 in its four cells,
sits just above the 0.5 threshold, so its $r=2$ and $r=4$ cells are ``collapse'' only narrowly. Moving the dormancy threshold
within 0.45--0.5 and the rank threshold within 50--70\% of the peak changes no cell; a dormancy threshold of
0.55 or 0.6 turns Ant at $r=2$ and $r=4$ into ``neither'' and changes nothing else.

\begin{table}[h]\centering\footnotesize
\caption{Collapse measures per phase-map cell (30k): median over runs of the effective-rank drop
$1-\text{final}/\text{peak}$ from the all-run peak and from the post-learning peak, the number of runs whose
peak precedes learning, and the median final dormant fraction (threshold 0). We count each run once, merging
the 7 replays at $r{=}4$ in each of Walker2d and Ant with their originals.}
\label{tab:collapse}
\setlength{\tabcolsep}{5pt}
\begin{tabular}{@{}lrrcccc@{}}
\toprule
 & & & \multicolumn{2}{c}{Rank drop from} & Peak before & Final \\
Cell & Runs & Flagged & all-run peak & post-learning peak & learning & dormant \\
\midrule
HalfCheetah $r{=}2$ & 8 & 0 & 0.59 & 0.06 & 8 & 0.77 \\
HalfCheetah $r{=}4$ & 8 & 0 & 0.65 & 0.10 & 8 & 0.77 \\
HalfCheetah $r{=}8$ & 8 & 0 & 0.68 & 0.06 & 8 & 0.77 \\
HalfCheetah $r{=}16$ & 32 & 0 & 0.69 & 0.05 & 32 & 0.78 \\
Walker2d $r{=}2$ & 8 & 0 & 0.72 & 0.14 & 8 & 0.82 \\
Walker2d $r{=}4$ & 32 & 0 & 0.74 & 0.24 & 32 & 0.82 \\
Walker2d $r{=}8$ & 8 & 8 & 0.80 & 0.33 & 8 & 0.82 \\
Walker2d $r{=}16$ & 10 & 10 & 0.85 & 0.43 & 10 & 0.82 \\
Ant $r{=}2$ & 6 & 0 & 0.87 & 0.61 & 6 & 0.51 \\
Ant $r{=}4$ & 32 & 0 & 0.97 & 0.90 & 32 & 0.51 \\
Ant $r{=}8$ & 6 & 3 & 0.99 & 0.96 & 6 & 0.52 \\
Ant $r{=}16$ & 8 & 8 & 1.00 & 0.95 & 8 & 0.51 \\
Hopper $r{=}2$ & 8 & 0 & 0.44 & 0.01 & 8 & 0.41 \\
Hopper $r{=}4$ & 8 & 0 & 0.25 & 0.01 & 8 & 0.41 \\
Hopper $r{=}8$ & 8 & 0 & 0.10 & 0.01 & 8 & 0.41 \\
Hopper $r{=}16$ & 10 & 1 & 0.17 & 0.01 & 10 & 0.41 \\
Humanoid $r{=}2$ & 8 & 0 & 0.01 & 0.01 & 0 & 0.47 \\
Humanoid $r{=}4$ & 8 & 0 & 0.05 & 0.05 & 0 & 0.47 \\
Humanoid $r{=}8$ & 8 & 0 & 0.07 & 0.04 & 2 & 0.58 \\
Humanoid $r{=}16$ & 10 & 0 & 0.19 & 0.17 & 2 & 0.70 \\
\bottomrule
\end{tabular}
\end{table}

\subsection{Resets}\label{app:reset} Table~\ref{tab:reset} gives the paired return comparison of
Figure~\ref{fig:reset} on its 6 paired seeds (its no-reset arm has 7 seeds and its reset arm 6; the no-reset
arm's median return at 60k is 425.5 over all 7 and 505.5 over the 6 paired ones). The no-reset arm's median return rises from
24.5 (mean over 20k--30k) to 425.5 at 60k. With 6 pairs, $p=0.031$ is the smallest value the exact Wilcoxon test
can give. Over 20k--60k the reset arm's lead is smaller and uncertain
(375.5 against 281.6, $p=0.22$), because return dips after each reset. The dormancy trigger fired at every
opportunity the 8k minimum gap allowed (at 6k, 14k, \dots, 54k), so the arm acts as a periodic reset. At 41k, 3k
steps after a reset, the no-reset arm is ahead in the median (Table~\ref{tab:reset}); snapshots depend on the phase of the reset
cycle, so window means are the fairer comparison. The low-UTD reference (HalfCheetah $r=4$, 12 seeds, same
trainer, width and batch, different seeds) has median return 592.3 at 60k and 402.1 over 50k--60k.

\begin{table}[h]\centering\small
\caption{HalfCheetah $r{=}16$: no reset against the dormancy-triggered reset, 6 paired seeds.}
\label{tab:reset}
\begin{tabular}{@{}lcccc@{}}
\toprule
Return measure & No reset (median) & Reset (median) & Reset ahead & Exact Wilcoxon $p$ \\
\midrule
Mean over 50k--60k & 491.0 & 876.6 & 6/6 & 0.031 \\
Mean over 20k--60k & 281.6 & 375.5 & 5/6 & 0.22 \\
At 60k & 505.5 & 1228.8 & 6/6 & 0.031 \\
Mean over 31k--41k & 239.5 & 126.7 & 2/6 & 0.56 \\
At 41k & 305.0 & 76.0 & 1/6 & 0.44 \\
\bottomrule
\end{tabular}
\end{table}

\section{Mechanism, Hazard and Gradient Clipping}\label{app:mech}
\begin{figure}[h]\centering\includegraphics[width=\linewidth]{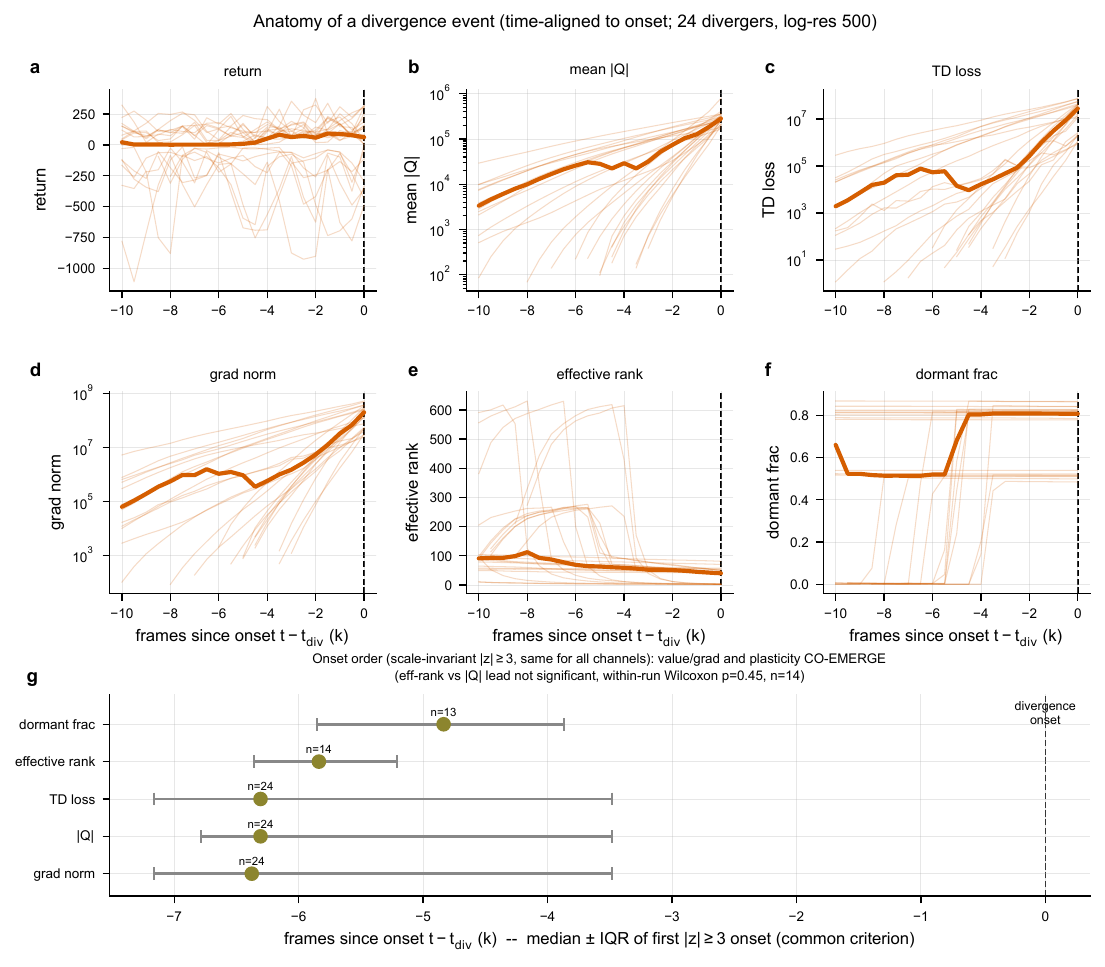}
\caption{24 flagged runs (Walker2d $r{=}8$ and $r{=}16$, Ant $r{=}16$; 8 each; logged every 500 steps)
aligned on their flag step. Terms in the panels: ``divergers'' are these flagged runs; $t_{\rm div}$,
``onset'' on the time axes and ``divergence onset'' denote the flag step; ``frames'' are environment steps;
``log-res 500'' is the 500-step logging interval. (a)--(f) Return, $\Q$, loss, gradient
norm, effective rank, dormant fraction; thin lines are runs, thick the median. (g) Median and interquartile
range of each channel's first $|z|\ge3$ departure (its ``onset'' in this panel); $n$ counts the runs in which the
channel departs. ``CO-EMERGE'' in its title means that the rank departs at about the same time as $\Q$: the
panel's Wilcoxon test, the one reported above, detects no lead either way.}\label{fig:S-anatomy}
\end{figure}

\paragraph{Event timing.} The event-timing runs replay seeds 0--7 of three phase-map cells with 500-step
logging. The onset of each channel is its first departure from its own early baseline (first quarter of the
10k steps before the flag) by $|z|\ge3$, the same rule for every channel (Figure~\ref{fig:S-anatomy}g). $\Q$
departs in all 24 runs (median 6.31k steps before the flag; since the baseline spans the first quarter of the
window, no onset can lead the flag by more than 7.5k). The effective rank departs in 14; across those 14
its onset trails that of $\Q$ by a median 0.5k steps (bootstrap 95\% CI $-0.5$k to $+1.0$k; mean 0.32k, 90\%
$t$ interval $-0.29$k to $+0.93$k; two-sided Wilcoxon $p=0.453$). The smallest tolerance within which a two
one-sided test declares the onsets equivalent is $\pm0.93$k, against a logging resolution of 0.5k; a $\pm2$k tolerance
gives $p=0.0002$, but we chose it after seeing the data. The analysis is conditional on the
14 of 24 runs in which rank departs at all; in the other 10, rank shows no such departure in the 10k steps
before the flag.

\begin{figure}[h]\centering\includegraphics[width=\linewidth]{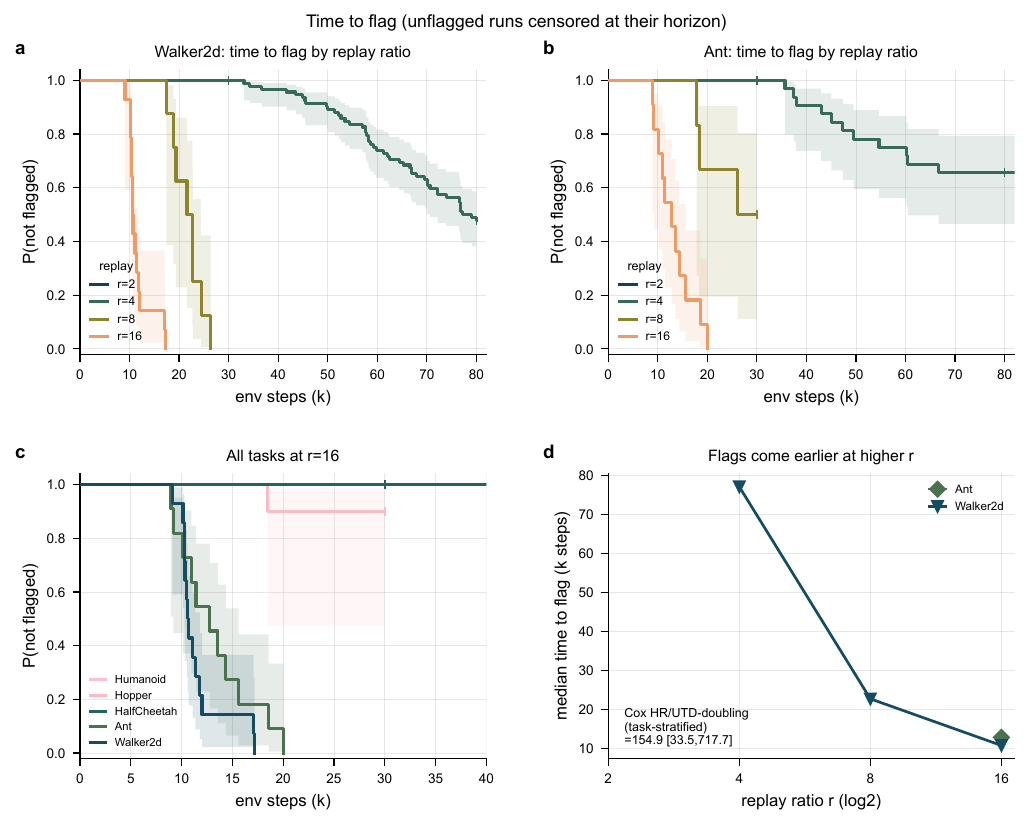}
\caption{Time to flag, with unflagged runs censored at their horizon, over the 427 unclipped SAC runs
outside the reset study and the controlled live arm (some of them repeat others; Appendix~\ref{app:compute}).
Kaplan--Meier curves with 95\% intervals by UTD ratio for Walker2d (a) and Ant (b), by task at $r{=}16$
(c), and the median time to flag against the UTD ratio with the task-stratified Cox hazard ratio
(d).}\label{fig:S-surv}
\end{figure}

\paragraph{Hazard.} We build Figure~\ref{fig:S-surv} from the 427 unclipped SAC runs outside the reset study (all three
of its arms) and the controlled live arm, dropping the replays and continuations that the rule of
Appendix~\ref{app:compute} detects. The 30k runs at $r=4$ and the runs to 200k enter alongside the runs they
repeat, so the 427 are not all distinct trajectories. We fit the task-stratified Cox model on the Walker2d and Ant runs among
them; its hazard ratio per doubling of the UTD ratio is 154.94 (95\% CI 33.45--717.65). This fit (also shown in panel
(d)) counts as independent runs the 64 runs to 30k at $r=4$ (32 Walker2d, 32 Ant), which repeat the first 30k
steps of longer runs with the same seeds, and its interval does not account for that repetition. The Schoenfeld-residual
check of the lifelines package finds no violation of proportional hazards at its default level 0.01; the log prints
its $p$ as nan because the check returns output only when it finds a violation.

\paragraph{Gradient clipping.} We clip the critic's gradient norm at 1 or 10 in Walker2d, Ant and Hopper at $r=16$,
with a 30k budget. Clipping leaves the failure in place: the guard still flags clipped Walker2d
and Ant runs, and on Walker2d at clip 1 it flags them at 10.5k--18.2k, against 9.1k--17.2k without clipping. The guard
also flags a clipped Hopper run, at 20.1k steps.

\section{TD3}\label{app:td3}
TD3 \citep{fujimoto2018td3} diverges too, at a higher UTD ratio than SAC. The guard flags no TD3 run at
$r\in\{4,8\}$ (0 of 16 runs at each ratio; Wilson upper bound 0.194), and it flags TD3 Walker2d runs at $r=16$.
There, on the 31 runs of the TD3 forecast cohort, the LOSO AUC of $g$ over the first 10k steps is 0.630
(95\% CI 0.386--0.859) with the SAC protocol, indistinguishable from chance, and the growth-abort controller
loses 81,365 steps (Table~\ref{tab:ledger}). We did not compare SAC and TD3 at a matched UTD ratio and window, so
this is no evidence that TD3 is less forecastable.

TD3 uses the same trainer, networks and guard. We compute $g$ over the records up to 10k steps
(the guard flags one run at 12,086). Adding TD3 Walker2d seeds 41--44 at $r=16$ (35 runs) gives a LOSO AUC of
0.611 (95\% CI 0.356--0.856). Table~\ref{tab:td3} gives the forecast under the SAC protocol (unweighted LOSO
logistic regression) and with class-balanced weights. No version differs clearly from chance.

\begin{table}[h]\centering\small
\caption{TD3 Walker2d $r{=}16$: LOSO AUC of $g$ (95\% bootstrap CI) on the forecast cohort. A window ending at 15k
drops the run flagged at 12k.}
\label{tab:td3}
\begin{tabular}{@{}llcc@{}}
\toprule
Runs & Window & SAC protocol & Balanced weights \\
\midrule
All 31 & $(5\text{k},10\text{k}]$ & 0.630 [0.386, 0.859] & 0.690 [0.462, 0.908] \\
30 & $(5\text{k},15\text{k}]$ & 0.534 [0.298, 0.764] & 0.652 [0.404, 0.870] \\
\bottomrule
\end{tabular}
\end{table}

\section{A DQN Pilot on MinAtar}\label{app:dqn}
A DQN \citep{mnih2015dqn} pilot on MinAtar Breakout \citep{young2019minatar} ran 199 runs (180 distinct trajectories, identical logs counted once) at UTD ratios 0.5 to 4, with a network of
width 128 or 1024 and 40k or 300k frames. No standard run crossed its guard ($\Q>10^4$ or loss $>10^6$): 0 of 172
distinct standard runs (Wilson upper bound 0.022), including 0 of 10 at UTD ratio 4 over 300k frames (0.278), and the
largest logged $\Q$ in any of them is 87.3. In most of these runs, all at UTD ratio 1, the target-copy interval exceeded
the run's length, so they never bootstrapped from a copied target. The only flags came in 8 engineered near-online runs (a
64-transition buffer, the target copied every step, a guard at $\Q>200$), all within 80 to 3,108 frames. In this
pilot, divergence was absent or immediate, with no window to forecast from. The agent, network and guard differ
from the SAC study, so this says nothing about whether the forecast transfers to discrete actions.

\section{Compute and Reproducibility}\label{app:compute}
All experiments\iflnsection{} except the LayerNorm pairs\fi{} ran on one node with 8 GPUs (48\,GB each; CUDA 13;
fp32 storage with TF32 matmuls), several runs per GPU, in eager mode: no run used \texttt{torch.compile}.\iflnsection{}
Each LayerNorm pair ran both arms on one GPU: pairs 800--807 on a workstation GPU, and pairs 808--831 on cluster
nodes with RTX A4500 or A5000 GPUs (Section~\ref{sec:ln}).\fi{}
The 612 runs took 1242.6 run-hours of wall-clock time in total. Software: Python 3.12, PyTorch, Gymnasium
\citep{towers2024gymnasium} and MuJoCo \citep{todorov2012mujoco}, scikit-learn, SciPy, NumPy. We tuned no
hyperparameter per task. Each run sets the seeds of Python, NumPy and PyTorch at its start, and on one machine, runs
with identical settings are deterministic given the seed: the continued controlled live runs match their uncontrolled twins at every
logged record.

\paragraph{Run census.} Table~\ref{tab:census} lists the TD3 and further characterization cohorts that
Table~\ref{tab:design} omits. The 612 runs also include the 7 runs of the rank-trigger reset arm
(Appendix~\ref{app:impl}), the 24 controlled live runs, 21 of them with an uncontrolled twin, the 14 replays of the phase map's $r=4$ cells
(below) and TD3 Walker2d seeds 41--44 at $r=16$ (Appendix~\ref{app:td3}).

\paragraph{Replays and continuations.} Of the 612 run records, 67 replay or continue another run with the same
task, UTD ratio and seed (identical $\Q$ at every common record, or, where the logging windows differ, the
same flag step): the 24 event-timing runs, 14 runs of the phase map's $r=4$ cells, the 8 Ant runs extended to
200k, and the 21 controlled live runs with a twin. That leaves 545 run records that this rule matches to no other.
The rule misses four groups. In three, the logging window or horizon differs from that of the run repeated and the
guard flags no run, so neither test applies: the 30k runs of the phase map's Walker2d and Ant cells at $r=4$ are the
first 30k steps of the 80k forecast runs with the same seeds (averaged over the same window, their $\Q$ agrees at
every common record); the HalfCheetah and Humanoid runs to 200k are phase-map seeds 0, 1, 5 and 6 at $r=16$ run on;
and the no-reset arm of the reset study runs the phase-map HalfCheetah seeds at $r=16$ on to 60k. The fourth is the
rank-trigger arm, whose trigger never fired, so it matches the no-reset arm at every record (Appendix~\ref{app:impl}). We count each run once in Figure~\ref{fig:phase} and
Table~\ref{tab:collapse}; Figure~\ref{fig:S-surv} counts the 30k runs at $r=4$ and the runs to 200k alongside the
runs they repeat. Every per-seed log (configuration, flag, and per-window return, $\Q$, loss,
gradient norm, effective rank and dormant fraction), the code, and the text output of every analysis
behind the numbers in this paper are available from the corresponding author on request.

\begin{table}[h]\centering\footnotesize
\caption{Cohorts that Table~\ref{tab:design} omits (SAC unless marked TD3). Runs: the runs we analyze; the
flagged counts of the event-timing and TD3 cohorts at $r\in\{4,8\}$ are in Appendices~\ref{app:mech}
and~\ref{app:td3}.}
\label{tab:census}
\setlength{\tabcolsep}{3pt}
\begin{tabular}{@{}lp{6.2cm}lcl@{}}
\toprule
Cohort & Tasks (UTD ratio $r$) & Runs & $H$ & Role \\
\midrule
Event timing & Walker2d ($r{=}8,16$), Ant ($r{=}16$), replays & 8 each & 40k & char. \\
Multi-task & Walker2d, Ant ($r{=}16$) & 4, 3 & 60k & char. \\
Gradient clipping & Walker2d, Ant, Hopper ($r{=}16$), clip 1 or 10 & 23 & 30k & char. \\
TD3 & Walker2d, Ant ($r{=}4,8$) & 8 each & 80k & char. \\
TD3 & Walker2d ($r{=}16$), forecast cohort & 31 & 80k & fit/eval \\
TD3 & Ant ($r{=}16$) & 5 & 80k & char. \\
\bottomrule
\end{tabular}
\end{table}

\iflnsection
\section{The LayerNorm Arm in Detail}\label{app:ln}
\paragraph{Collapse symptoms and the rate $g$.} The LayerNorm critic ends with its units active and its rank high.
The largest batch-mean $\Q$ a run reaches has median $5.19\times10^3$ with LayerNorm against $1.82\times10^5$ without
(Figure~\ref{fig:ln}a). The last logged dormant fraction has median 0.000 with LayerNorm against 0.829, and the
post-learning rank drop 0.052 against 0.245 (Figure~\ref{fig:ln}b shows the rank). The early-growth signal $g$ is lower
with LayerNorm in 31 of the 32 pairs (median 0.1207 against 0.1600 per 1k steps; paired difference $-0.0419$, 95\%
CI $-0.0553$ to $-0.0310$; Wilcoxon $p=9.3\times10^{-10}$). LayerNorm thus changes the rank drop and dormancy along
with the rate $g$ and the flag, so this arm cannot attribute the flag to $\Q$ growth or to collapse. The
HalfCheetah critics of Section~\ref{sec:decouple}, which collapse while $\Q$ stays of order $10^2$, separate $\Q$
growth from dormancy and from the rank drop at learning onset. They show little post-learning rank drop (0.05--0.10,
Table~\ref{tab:collapse}), and that drop co-emerges with $\Q$ growth, so neither this arm nor HalfCheetah
separates it from $\Q$ growth.

\paragraph{Forecast inside each arm.} With no flagged LayerNorm run, the early-growth signal has no event to forecast
in that arm. In the plain arm, $g$ ranks the 17 flagged runs above the 15 unflagged ones with AUC 0.714 (95\% CI
0.518--0.875; LOSO AUC 0.651), and the flagged runs have median $g$ 0.1723 against 0.1591 for the unflagged ones. The plain-critic reference for the forecast
is the cohort of Section~\ref{sec:forecast} (Table~\ref{tab:forecast}).

\paragraph{Plain arm, return and scope.} The plain arm's flag rate is consistent with that of the 64 seeds of
Section~\ref{sec:forecast} (Fisher $p=0.83$). On the 15 pairs in which the guard flags neither arm, the LayerNorm
critic's mean return over 70k--80k has median 84.8 against 251.7 for the plain critic (Wilcoxon $p=0.0015$); over all
32 LayerNorm runs it has median 90.4. The result covers $r=4$ within 80k steps on this network, and it shows that
LayerNorm acts on the quantity this paper forecasts. LayerNorm at higher UTD ratios is the next test
(Section~\ref{sec:limits}).
\fi

\end{document}